\documentclass[Afour,sageh,times]{sagej}

\usepackage{moreverb,url}
\usepackage{multicol}

\usepackage[colorlinks,bookmarksopen,bookmarksnumbered,citecolor=red,urlcolor=red]{hyperref}

\newcommand\BibTeX{{\rmfamily B\kern-.05em \textsc{i\kern-.025em b}\kern-.08em
T\kern-.1667em\lower.7ex\hbox{E}\kern-.125emX}}

\usepackage{graphics} % for pdf, bitmapped graphics files
\usepackage{epsfig} % for postscript graphics files
\usepackage{mathptmx} % assumes new font selection scheme installed
\usepackage{times} % assumes new font selection scheme installed
\usepackage{amsmath} % assumes amsmath package installed
\usepackage{amssymb}  % assumes amsmath package installed
\usepackage{siunitx} % for SI units
\usepackage{textcomp}
\usepackage{gensymb} % for degree command
\usepackage{booktabs}
\usepackage{multirow}
\usepackage{xcolor}

\let\labelindent\relax
\usepackage{enumitem}
\usepackage{xspace}

\usepackage{makecell}
\usepackage{multicol}
\usepackage{rotating}
\usepackage[percent]{overpic}
\usepackage{contour}
\usepackage{courier}

\usepackage{subcaption} % issues a warning with CVPR/ICCV format
\usepackage[font=small]{caption}
\usepackage[percent]{overpic}

\usepackage[11pt]{moresize}

\usepackage{pifont}% http://ctan.org/pkg/pifont
\definecolor{MyDarkBlue}{rgb}{0,0.08,1}
\definecolor{airforceblue}{rgb}{0.36, 0.54, 0.66}
\definecolor{MyDarkGreen}{rgb}{0.02,0.6,0.02}
\definecolor{MyDarkRed}{rgb}{0.8,0.02,0.02}
\definecolor{MyDarkOrange}{rgb}{0.40,0.2,0.02}
\definecolor{MyPurple}{RGB}{111,0,255}
\definecolor{MyRed}{rgb}{1.0,0.0,0.0}
\definecolor{MyGold}{rgb}{0.75,0.6,0.12}
\definecolor{MyDarkgray}{rgb}{0.66, 0.66, 0.66}
\definecolor{MyPink}{rgb}{0.9, 0.33, 0.5}
\definecolor{MyCyan}{rgb}{0., 0.4, 0.4}

\usepackage{tikz}

\newcommand{\legalTM}{\textsuperscript{\texttrademark}}

\usepackage{tikz}

\begin{document}

\runninghead{Diffusion Policy}

\title{Diffusion Policy: Visuomotor Policy Learning via Action Diffusion}
% \title{.}

\author{Cheng Chi$^{*}$\affilnum{1}, Zhenjia Xu$^{*}$\affilnum{1}, Siyuan Feng\affilnum{2}, Eric Cousineau\affilnum{2}, Yilun Du\affilnum{3},  Benjamin Burchfiel\affilnum{2}, Russ Tedrake \affilnum{2,3}, Shuran Song\affilnum{1,4}}

\affiliation{\affilnum{*} Joint First Author \\
\affilnum{1}Columbia University, US
\affilnum{2}Toyota Research Institute, US.
\affilnum{3}MIT, US.
\affilnum{4}Stanford University, US.
}

\corrauth{Cheng Chi, Columbia University, US}
\email{chenng.chi@columbia.edu}

\begin{abstract} This paper introduces Diffusion Policy, a new way of generating robot behavior by representing a robot's visuomotor policy as a conditional denoising diffusion process. We benchmark Diffusion Policy across 15 different tasks from 4 different robot manipulation benchmarks and find that it consistently outperforms existing state-of-the-art robot learning methods with an average improvement of 46.9\%. 
Diffusion Policy learns the gradient of the action-distribution score function and iteratively optimizes with respect to this gradient field during inference via a series of stochastic Langevin dynamics steps.
We find that the diffusion formulation yields powerful advantages when used for robot policies, including gracefully handling multimodal action distributions, being suitable for high-dimensional action spaces, and exhibiting impressive training stability.
To fully unlock the potential of diffusion models for visuomotor policy learning on physical robots, this paper presents a set of key technical contributions including the incorporation of receding horizon control, visual conditioning, and the time-series diffusion transformer. 
We hope this work will help motivate a new generation of policy learning techniques that are able to leverage the powerful generative modeling capabilities of diffusion models. Code, data, and training details is available \url{diffusion-policy.cs.columbia.edu}
\end{abstract}
% Cover letter
% https://docs.google.com/document/d/1culy_CpPnacKd7t0tIkUwMVL6aU-bS-yiKFrnbBn2CQ/edit

\twocolumn[{
\renewcommand\twocolumn[1][]{#1}
\maketitle
    \vspace{-5mm}
\begin{center}
    \includegraphics[width=0.95\textwidth]{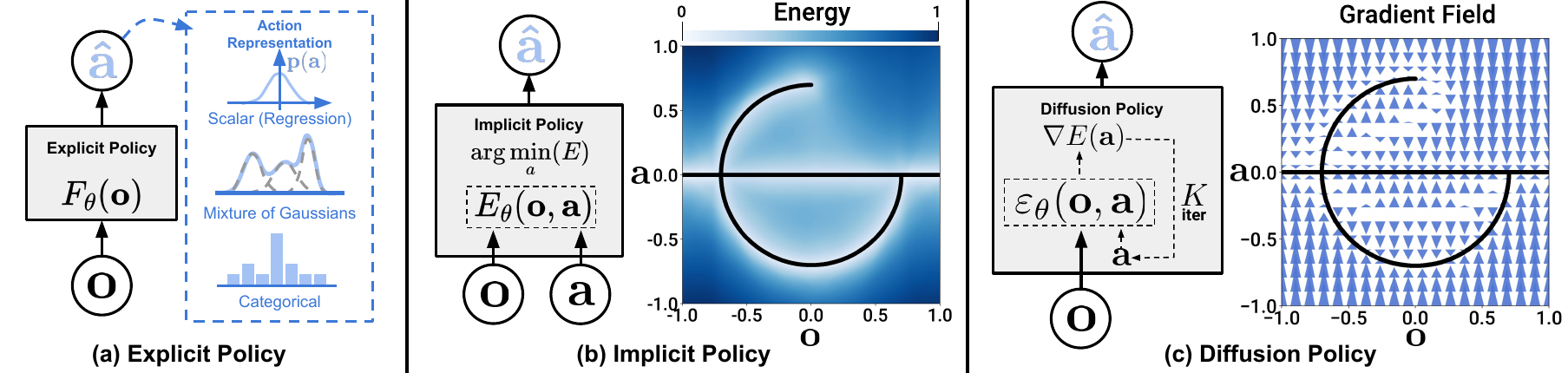}
    \captionof{figure}{\textbf{Policy Representations.} \label{fig:policy_rep} a) Explicit policy with different types of action representations.  b) Implicit policy learns an energy function conditioned on both action and observation and optimizes for actions that minimize the energy landscape c) Diffusion policy refines noise into actions via a learned gradient field. This formulation provides stable training, allows the learned policy to accurately model multimodal action distributions, and accommodates high-dimensional action sequences. 
    % \todo{update figure to make b and c consistent, both 2D or both 3D. Make it clear c is gradient of b change J (a) -> E (a)}
    } 
%https://docs.google.com/drawings/d/1SNd5_khk3RsYuE9JCwUVjmRED-eF3UrO78XnzbOwE4Y/edit?usp=sharing
\end{center}
}]

% \begin{figure*}[!h]
% \centering
% \includegraphics[width=\linewidth]{figure/DP_teaser.pdf}
% % https://docs.google.com/drawings/d/1SNd5_khk3RsYuE9JCwUVjmRED-eF3UrO78XnzbOwE4Y/edit?usp=sharing
% \caption{\textbf{Policy Representations.} \label{fig:policy_rep} a) Explicit policy with different types of action representations.  b) Implicit policy learns an energy function conditioned on both action and observation and optimizes for actions that minimize the energy landscape c) Diffusion policy refines noise into actions via a learned gradient field. This formulation provides stable training, allows the learned policy to accurately model multimodal action distributions, and accommodates high-dimensional action sequences.} 
% % \label{fig:my_label}
% \end{figure*}

\section{Introduction}
% Policy learning, in its simplest form, can be treated as a supervised regression task with actions learned via a regression loss.
Policy learning from demonstration, in its simplest form, can be formulated as the supervised regression task of learning to map observations to actions.
In practice however, the unique nature of predicting robot actions --- such as the existence of multimodal distributions, sequential correlation, and the requirement of high precision ---  makes this task distinct and challenging compared to other supervised learning problems.   

%Behavior cloning is a promising method for creating complex and highly reactive robot behaviors. This approach formulates behavior learning as a supervised learning problem where a policy is conditioned on observations and taught to regress expert-demonstrated actions. Despite this simple problem formulation, the unique nature of predicting robot actions --- such as the presence of multi-modality, sequential correlation, and the requirement of high precision --- makes this task distinct and challenging compared to other supervised learning problems and straightforwardly applying existing approaches from image or video-based regression have shown poor results \cite{robomimic}.  

%input representations \todo{needs citations} -- such as explicit state, multiple viewpoints, temporal history, and frozen pretrained feature extractors --
Prior work attempts to address this challenge by exploring different \textit{action representations} (Fig \ref{fig:policy_rep} a) -- using mixtures of Gaussians \cite{robomimic}, categorical representations of quantized actions \cite{bet},
or by switching the \textit{the policy representation} (Fig \ref{fig:policy_rep} b) -- from  explicit to implicit to better capture multi-modal distributions \cite{ibc,wu2020spatial}.

In this work, we seek to address this challenge by introducing a new form of robot visuomotor policy that generates behavior via a ``conditional denoising diffusion process \cite{ho2020denoising} on robot action space'', \textbf{Diffusion Policy}. In this formulation, instead of directly outputting an action, the policy infers the action-score gradient, conditioned on visual observations, for $K$ denoising iterations (Fig. \ref{fig:policy_rep} c). 
This formulation allows robot policies to inherit several key properties from diffusion models -- significantly improving performance.
\begin{itemize} %[leftmargin=3mm]
    
    \item \textbf{Expressing multimodal action distributions.} 
     By learning the gradient of the action score function \cite{song2019score} and performing Stochastic Langevin Dynamics sampling on this gradient field, Diffusion policy can express arbitrary normalizable distributions \cite{neal2011mcmc}, which includes multimodal action distributions, a well-known challenge for policy learning.
     %This capability also enables better compatibility with positional action commands, which is typically multimodal but necessary for high-precision tasks. 

    \item \textbf{High-dimensional output space.} As demonstrated by their impressive image generation results, diffusion models have shown excellent scalability to high-dimension output spaces. This property allows the policy to jointly infer a \textit{sequence} of future actions instead of \textit{single-step} actions, which is critical for encouraging temporal action consistency and avoiding myopic planning.  
    
    \item \textbf{Stable training.} 
    Training energy-based policies often requires negative sampling to estimate an intractable normalization constant, which is known to cause training instability \cite{du2020improved,ibc}. Diffusion Policy bypasses this requirement by learning the gradient of the energy function and thereby achieves stable training while maintaining distributional expressivity. %This training stability is crucial when selecting the optimal checkpoint requires expensive physical robot executions.

    %\item Data efficiency(?) Is it true? Not sure why 
\end{itemize}

Our \textbf{primary contribution} is to bring the above advantages to the field of robotics and demonstrate their effectiveness on complex real-world robot manipulation tasks. To successfully employ diffusion models for visuomotor policy learning, we present the following technical contributions that enhance the performance of Diffusion Policy and unlock its full potential on physical robots:
\begin{itemize} %[leftmargin=3mm]
    \item \textbf{Closed-loop action sequences.} We combine the policy's capability to predict high-dimensional action sequences with \textit{receding-horizon control} to achieve robust execution. This design allows the policy to continuously re-plan its action in a closed-loop manner while maintaining temporal action consistency -- achieving a balance between long-horizon planning and responsiveness. 

    %\item We introduced end-to-end trained real-time vision input to diffusion policy. By treating observations as conditioning instead of being part of the joint data distribution to be modeled by DDPM, diffusion policy executes the vision encoder once regardless of the number of denoising iterations, drastically reducing required computation power and inference latency, enabling the use of DDPM in real-time visuomotor policies.

    \item \textbf{Visual conditioning.} We introduce a vision-conditioned diffusion policy, where the visual observations are treated as conditioning instead of a part of the joint data distribution.  In this formulation, the policy extracts the visual representation once regardless of the denoising iterations, which drastically reduces the computation and enables real-time action inference.

    \item \textbf{Time-series diffusion transformer.} We propose a new transformer-based diffusion network  that minimizes the over-smoothing effects of typical CNN-based models and achieves state-of-the-art performance on tasks that require high-frequency action changes and velocity control. 
\end{itemize}

We systematically evaluate Diffusion Policy across \textbf{15} tasks from \textbf{4} different benchmarks \cite{ibc, gupta2019relay, robomimic, bet} under the behavior cloning formulation. The evaluation includes both simulated and real-world environments, 2DoF to 6DoF actions, single- and multi-task benchmarks, and fully- and under-actuated systems, with rigid and fluid objects, using demonstration data collected by single and multiple users.  

Empirically, we find \textbf{consistent} performance boost across all benchmarks with an average improvement of 46.9\%, providing strong evidence of the effectiveness of Diffusion Policy. We also provide detailed analysis to carefully examine the characteristics of the proposed algorithm and the impacts of the key design decisions. 

This work is an extended version of the conference paper \cite{chi2023diffusionpolicy}.  We expand the content of this paper in the following ways: 
\begin{itemize} %[leftmargin=3mm]
\item Include a new discussion section on the connections between diffusion policy and control theory. See Sec. \ref{sec:control}.   
\item Include additional ablation studies in simulation on alternative network architecture design and different pretraining and finetuning paradigms, Sec. \ref{sec:arch_ablation}.  
\item Extend the real-world experimental results with three bimanual manipulation tasks including  Egg Beater, Mat Unrolling, and Shirt Folding in  Sec. \ref{sec:eval_bimanual}. 
\end{itemize}

The code, data, and training details are publicly available for reproducing our results \url{diffusion-policy.cs.columbia.edu}.  
%\textit{\textbf{Please refer to the supplementary material for the robot videos.}}
% \begin{figure}[t]
% \centering
% \includegraphics[width=0.98\linewidth]{figure/real_tasks_teaser.pdf}
% % https://docs.google.com/drawings/d/1j4ZPASr2r3dAHBx6qXbJUYoDzBZr-iOzIce6SWfzRCQ/edit
% \caption{\textbf{Realworld Benchmarks.} We deployed Diffusion Policy on two different robot platforms (UR5 and Franka) for 4 challenging tasks: under-actuate precise pushing (Push-T), 6DoF mug flipping, 6DoF sauce pouring, and periodic sauce spreading.  Please check out the supplementary material for the resulting robot videos. }
% % \label{fig:my_label}

% \end{figure}

\begin{figure*}[t]
    \centering
    \includegraphics[width=\linewidth]{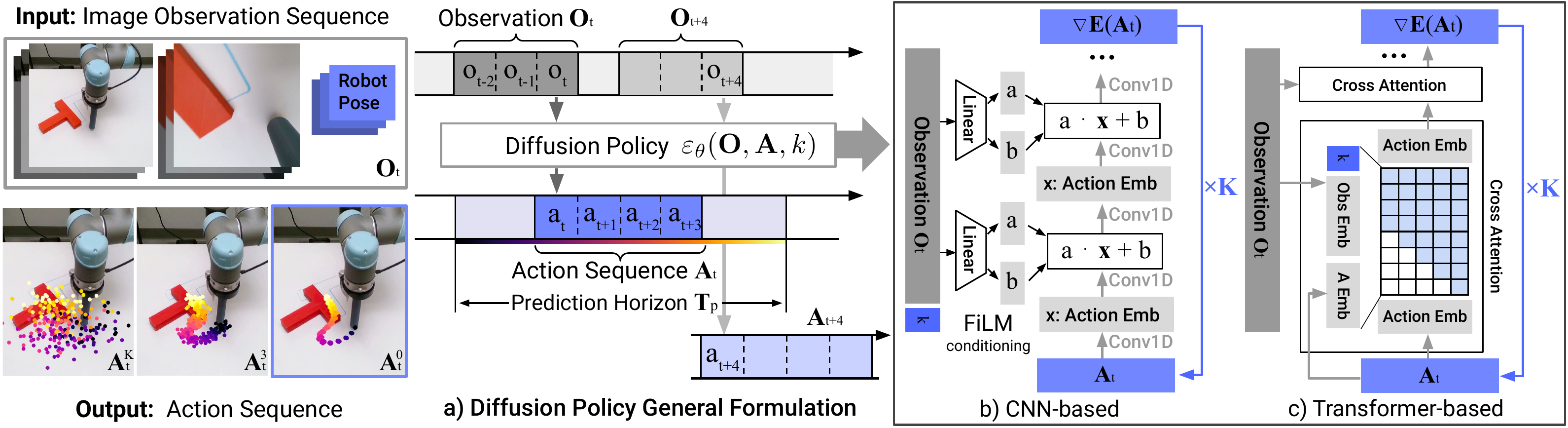}
    % https://docs.google.com/drawings/d/1Z-OWGff7cpdeAJ5V07L2IxZSEB-P0DDHW8zRXLku4sg/edit
    \caption{\textbf{Diffusion Policy Overview} \label{fig:policy_io} a) General formulation. At time step $t$, the policy takes the latest $T_o$ steps of observation data $O_t$ as input and outputs $T_a$ steps of actions $A_t$.  b) In the CNN-based Diffusion Policy, FiLM (Feature-wise Linear Modulation) \cite{perez2018film} conditioning of the observation feature $O_t$ is applied to every convolution layer, channel-wise. Starting from $\mathbf{A}^K_t$ drawn from Gaussian noise, the output of noise-prediction network $\epsilon_\theta$ is subtracted, repeating $K$ times to get $\mathbf{A}^0_t$, the denoised action sequence. c) In the Transformer-based \cite{vaswani2017attention} Diffusion Policy, the embedding of observation $\mathbf{O}_t$ is passed into a multi-head cross-attention layer of each transformer decoder block. Each action embedding is constrained to only attend to itself and previous action embeddings (causal attention) using the attention mask illustrated.  }
\end{figure*}

\section{Diffusion Policy Formulation}
\label{sec:method}

We formulate visuomotor robot policies as Denoising Diffusion Probabilistic Models (DDPMs) \cite{ho2020denoising}. Crucially, Diffusion policies are able to express complex multimodal action distributions and possess stable training behavior -- requiring little task-specific hyperparameter tuning.
% We represent visuomotor policies as Denoising Diffusion Probabilistic Models (DDPM) to leverage their ability to express complex multimodal distributions, especially in comparison with exponential family distributions such as Gaussian (L2 regression), Mixture of Gaussian (GMM) and Categorical (classification). 
% This added representation power is vital to Diffusion Policy's improved performance over the existing method and will be further discussed in Sec \ref{sec:multimodal}.
The following sections describe DDPMs in more detail and explain how they may be adapted to represent visuomotor policies. 

\subsection{Denoising Diffusion Probabilistic Models} 
\label{sec:ddpm}
DDPMs are a class of generative model where the output generation is modeled as a denoising process, often called Stochastic Langevin Dynamics \cite{welling2011bayesian}. 

Starting from $\mathbf{x}^K$ sampled from Gaussian noise, the DDPM performs $K$ iterations of denoising to produce a series of intermediate actions with decreasing levels of noise,
$\mathbf{x}^k, \mathbf{x}^{k-1} ...\mathbf{x}^{0} $, until a desired noise-free output $\mathbf{x}^0$ is formed.  
The process follows the equation
\vspace{-1mm}
\begin{equation}
    \textbf{x}^{k-1}=\alpha(\textbf{x}^{k}-\gamma\epsilon_\theta(\mathbf{x}^k,k) + \mathcal{N} \bigl(0, \sigma^2 I \bigl)),
    \label{eq:unconditional_langevin}
\vspace{-1mm}
\end{equation}
where $\epsilon_\theta$ is the noise prediction network with parameters $\theta$ that will be optimized through learning and $\mathcal{N} \bigl(0, \sigma^2 I \bigl)$ is Gaussian noise added at each iteration.

The above equation \ref{eq:unconditional_langevin} may also be interpreted as a single noisy gradient descent step: 
\vspace{-2mm}
\begin{equation}
    \mathbf{x}'=\mathbf{x}-\gamma\nabla E(\mathbf{x}),
    \label{eq:gradient_descent}
\vspace{-1mm}
\end{equation}
where the noise prediction network $\epsilon_\theta(\mathbf{x},k)$ effectively predicts the gradient field $\nabla E(\mathbf{x})$, and  $\gamma$ is the learning rate. 

The choice of $\alpha,\gamma,\sigma$ as functions of iteration step $k$, also called noise schedule, can be interpreted as learning rate scheduling in gradient decent process. 
An $\alpha$ slightly smaller than $1$ has been shown to improve stability \cite{ho2020denoising}.
Details about noise schedule will be discussed in Sec \ref{sec:method-noise-schedule}.
% weights the importance of different frequencies of information by controlling the ''step size`` and the amount added noise for each iteration. They can be interpreted as learning rate scheduling in gradient decent process. 
% \cheng{Where should we discuss what kind of noise schedule we used?}

% The choice of $\alpha,\gamma,\sigma$ controls the schedule of added noise with respect to iteration step $k$. They are constant with respect to $\mathbf{x}$, and can be interpreted as learning rate scheduling in gradient decent process. 

\subsection{DDPM Training} 
\label{sec:ddpm_inference}

The training process starts by randomly drawing unmodified examples, $\mathbf{x}^0$, from the dataset. For each sample, we randomly select a denoising iteration $k$ and then sample a random noise $\mathbf{\epsilon}^k$ with appropriate variance for iteration $k$. The noise prediction network is asked to predict the noise from the data sample with noise added.

% Training the noise prediction network $\epsilon_\theta(\mathbf{x}^0,k)$ starts by randomly drawing unmodified data $\mathbf{x}^0$ from the dataset. For each data sample, we randomly select a denoising iteration $k$ and then sample a random noise $\mathbf{\epsilon}^k$ with appropriate variance for iteration $k$. The noise prediction network is asked to predict the noise from the data sample with noise added.
\vspace{-2mm}
\begin{equation}
    \mathcal{L} = MSE(\mathbf{\epsilon}^k, \epsilon_\theta(\mathbf{x}^0+\mathbf{\epsilon}^k,k))
    \label{eq:unconditional_loss}
\end{equation}

As shown in \cite{ho2020denoising}, minimizing the loss function in Eq \ref{eq:unconditional_loss} also minimizes the variational lower bound of the KL-divergence between the data distribution $p(\mathbf{x}^0)$ and the distribution of samples drawn from the DDPM $q(\mathbf{x}^0)$ using Eq \ref{eq:unconditional_langevin}.

\subsection{Diffusion for Visuomotor Policy Learning} 

While DDPMs are typically used for image generation ($\mathbf{x}$ is an image), we use a DDPM to learn robot visuomotor policies. This requires two major modifications in the formulation: 
1. changing the output $\mathbf{x}$  to represent robot actions.  
2. making the denoising processes \textit{conditioned} on input observation $\mathbf{O}_t$. 
The following paragraphs discuss each of the modifications, and Fig. \ref{fig:policy_io} shows an overview. 

\textbf{Closed-loop action-sequence prediction:}
%motivation, what we want
An effective action formulation should encourage temporal consistency and smoothness in long-horizon planning while allowing prompt reactions to unexpected observations. 
% idea of a solution 
% To accomplish this goal, we integrate the action-sequence prediction produced by a diffusion model with receding horizon control \cite{mayne1988receding} to achieve robust action execution. 
To accomplish this goal, we commit to the action-sequence prediction produced by a diffusion model for a fixed duration before replanning. 
%how we do it.
Concretely, at time step $t$ the policy takes the latest $T_o$ steps of observation data $\mathbf{O}_t$ as input and predicts $T_p$ steps of actions, of which $T_a$ steps of actions are executed on the robot without re-planning. Here, we define $T_o$ as the observation horizon, $T_p$ as the action prediction horizon and $T_a$ as the action execution horizon. 
% After policy prediction, the robot executes the $T_a$ steps of actions without re-planning, $T_a$ is the action horizon. The policy infers a new action sequence at the $t+T_a$ step. 
% point out the difference again
% Unlike other approaches \cite{robomimic,ibc, bet,janner2022diffuser}, we choose $T_a$ to be a value between $1$ and the full demonstration sequence length. 
% advantage 
This encourages temporal action consistency while remaining responsive. More details about the effects of $T_a$ are discussed in Sec \ref{sec:action_sequence}.
Our formulation also allows receding horizon control \cite{mayne1988receding} to futher improve action smoothness by warm-starting the next inference setup with previous action sequence prediction.
 % \shuran{@cheng, double check prediction horizon and action horizon}

\textbf{Visual observation conditioning:}
% We use DDPM to approximate the conditional distribution $p(\mathbf{A}_t |\mathbf{O}_t)$, instead of the joint distribution for both observation and action sequences $p(\mathbf{A}_t,\mathbf{O}_t)$ as done in planning applications \cite{janner2022diffuser}. 
% This allows the policy to infer action conditioned on observation but not at cost of generating future states — which will drastically slow down the diffusion process and decrease the accuracy of generated actions.
We use a DDPM to approximate the conditional distribution $p(\mathbf{A}_t | \mathbf{O}_t)$ instead of the joint distribution $p(\mathbf{A}_t,\mathbf{O}_t)$ used in \citet{janner2022diffuser} for planning. This formulation allows the model to predict actions conditioned on observations without the cost of inferring future states, speeding up the diffusion process and improving the accuracy of generated actions.
To capture the conditional distribution $p(\mathbf{A}_t |\mathbf{O}_t)$, we modify Eq \ref{eq:unconditional_langevin} to:
% \vspace{-0.5mm}
\begin{equation}
    \label{eq:diffusion_policy_langevin}
    \mathbf{A}^{k-1}_t = \alpha(\mathbf{A}^k_t - \gamma\epsilon_\theta(\mathbf{O}_t,\mathbf{A}^k_t,k) + \mathcal{N} \bigl(0, \sigma^2 I \bigl))
% \vspace{-1mm}
\end{equation}
The training loss is modified from Eq \ref{eq:unconditional_loss} to:
% The exclusion of observation features $\mathbf{O}_t$ from the denoising process also allows us to train the vision encoder \textbf{end-to-end} with the noise prediction networking using the training loss modified from Eq \ref{eq:unconditional_loss} to
% \vspace{-0.5mm}
\begin{equation}
    \label{eq:diffusion_policy_loss}
    \mathcal{L}=MSE(\mathbf{\epsilon}^k,\epsilon_\theta(\mathbf{O}_t, \mathbf{A}^0_t + \mathbf{\epsilon}^k, k))
% \vspace{-1mm}
\end{equation}

The exclusion of observation features $\mathbf{O}_t$ from the output of the denoising process significantly improves inference speed and better accommodates real-time control. It also helps to make \textbf{end-to-end} training of the vision encoder feasible.
Details about the visual encoder are described in Sec. \ref{sec:method-visual}.

\section{Key Design Decisions}
 In this section, we describe key design decisions for Diffusion Policy as well as its concrete implementation of $\epsilon_\theta$ with neural network architectures.
 
\subsection{Network Architecture Options}
\label{sec:method-network}
The first design decision is the choice of neural network architectures for $\epsilon_\theta$. 
In this work, we examine two common network architecture types, convolutional neural networks (CNNs) \cite{ronneberger2015u} and Transformers \cite{vaswani2017attention}, and compare their performance and training characteristics. Note that the choice of noise prediction network $\epsilon_\theta$ is independent of visual encoders, which will be described in Sec. \ref{sec:method-visual}.

\textbf{CNN-based Diffusion Policy} 
% \shuran{just shorten this part, the difference is already highlighted earlier, here just talk about how CNN is implemented}
We adopt the 1D temporal CNN from \citet{pmlr-v162-janner22a} with a few modifications:
First, we only model the conditional distribution $p(\mathbf{A}_t|\mathbf{O}_t)$ by conditioning the action generation process on observation features $\mathbf{O}_t$ with Feature-wise Linear Modulation (FiLM) \cite{perez2018film} as well as denoising iteration $k$, shown in Fig \ref{fig:policy_io} (b).
Second, we only predict the action trajectory instead of the concatenated observation action trajectory. 
Third, we removed inpainting-based goal state conditioning due to incompatibility with our framework utilizing a receding prediction horizon. However, goal conditioning is still possible with the same FiLM conditioning method used for observations.

In practice, we found the CNN-based backbone to work well on most tasks out of the box without the need for much hyperparameter tuning. However, it performs poorly when the desired action sequence changes quickly and sharply through time (such as velocity command action space), likely due to the inductive bias of temporal convolutions to prefer low-frequency signals \cite{tancik2020fourier}. 

\textbf{Time-series diffusion transformer}
% \shuran{use a consistent name for this. It is a contribution of the paper. }
To reduce the over-smoothing effect in CNN models \cite{tancik2020fourier}, we introduce a novel transformer-based DDPM which adopts the transformer architecture from minGPT \cite{bet} for action prediction. 
Actions with noise $A_t^k$ are passed in as input tokens for the transformer decoder blocks, with the sinusoidal embedding for diffusion iteration $k$ prepended as the first token. 
The observation $\mathbf{O}_t$ is transformed into observation embedding sequence by a shared MLP, which is then passed into the transformer decoder stack as input features.
% The observation $O_t$ is passed into the transformer encoder stack, with the transformer blocks replaced by a single linear layer. 
The ``gradient" $\epsilon_\theta(\mathbf{O_t},\mathbf{A_t}^k,k)$ is predicted by each corresponding output token of the decoder stack. 

In our state-based experiments, most of the best-performing policies are achieved with the transformer backbone, especially when the task complexity and rate of action change are high. However, we found the transformer to be more sensitive to hyperparameters. The difficulty of transformer training \cite{liu2020understanding} is not unique to Diffusion Policy and could potentially be resolved in the future with improved transformer training techniques or increased data scale.

\textbf{Recommendations.} 
 In general, we recommend starting with the CNN-based diffusion policy implementation as the first attempt at a new task. If performance is low due to task complexity or high-rate action changes, then the Time-series Diffusion Transformer formulation can be used to potentially improve performance at the cost of additional tuning. 

\subsection{Visual Encoder}
\label{sec:method-visual}
The visual encoder  maps the raw image sequence into a latent embedding $O_t$ and is trained end-to-end with the diffusion policy. 
Different camera views use separate encoders, and images in each timestep are encoded independently and then concatenated to form $O_t$.  
We used a standard ResNet-18 (without pretraining) as the encoder with the following modifications: 
1) Replace the global average pooling with a spatial softmax pooling to maintain spatial information \cite{robomimic}. 
2) Replace BatchNorm with GroupNorm \cite{groupnorm} for stable training. This is important when the normalization layer is used in conjunction with Exponential Moving Average \cite{he2020moco} (commonly used in DDPMs). 

\subsection{Noise Schedule}
\label{sec:method-noise-schedule}
The noise schedule, defined by $\sigma$, $\alpha$, $\gamma$ and the additive Gaussian Noise $\epsilon^k$ as functions of $k$, has been actively studied \cite{ho2020denoising, nichol2021improved}.  The underlying noise schedule controls the extent to which diffusion policy captures high and low-frequency characteristics of action signals. In our control tasks, we empirically found that the Square Cosine Schedule proposed in iDDPM \cite{nichol2021improved} works best for our tasks. 

\subsection{Accelerating Inference for Real-time Control}
We use the diffusion process as the policy for robots; hence, it is critical to have a fast inference speed for closed-loop real-time control. The Denoising Diffusion Implicit Models (DDIM) approach \cite{song2021ddim}  decouples the number of denoising iterations in training and inference, thereby allowing the algorithm to use fewer iterations for inference to speed up the process. In our real-world experiments, using DDIM with 100 training iterations and 10 inference iterations enables 0.1s inference latency on a Nvidia 3080 GPU.

\section{Intriguing Properties of Diffusion Policy}
In this section, we provide some insights and intuitions about diffusion policy and its advantages over other forms of policy representations.

\subsection{Model Multi-Modal Action Distributions}
\label{sec:multimodal}
The challenge of modeling multi-modal distribution in human demonstrations has been widely discussed in behavior cloning literature \cite{ibc,bet,robomimic}. Diffusion Policy's ability to express multimodal distributions naturally and precisely is one of its key advantages. 

%As described in Sec \ref{sec:ddpm} and illustrated in \ref{fig:policy_rep}, diffusion policy learns to predict the underlying score of the gradient field of the action distribution through denoising. Inference on diffusion policy then corresponds to a Stochastic Langevin Dynamics sampling procedure on this gradient field. %, which is probably able to draw samples from all modes in an action distribution $p(\mathbf{A}_t|\mathbf{O}_t$) \citep{neal2011mcmc}.

\begin{figure}[h]
\centering
\includegraphics[width=0.98\linewidth]{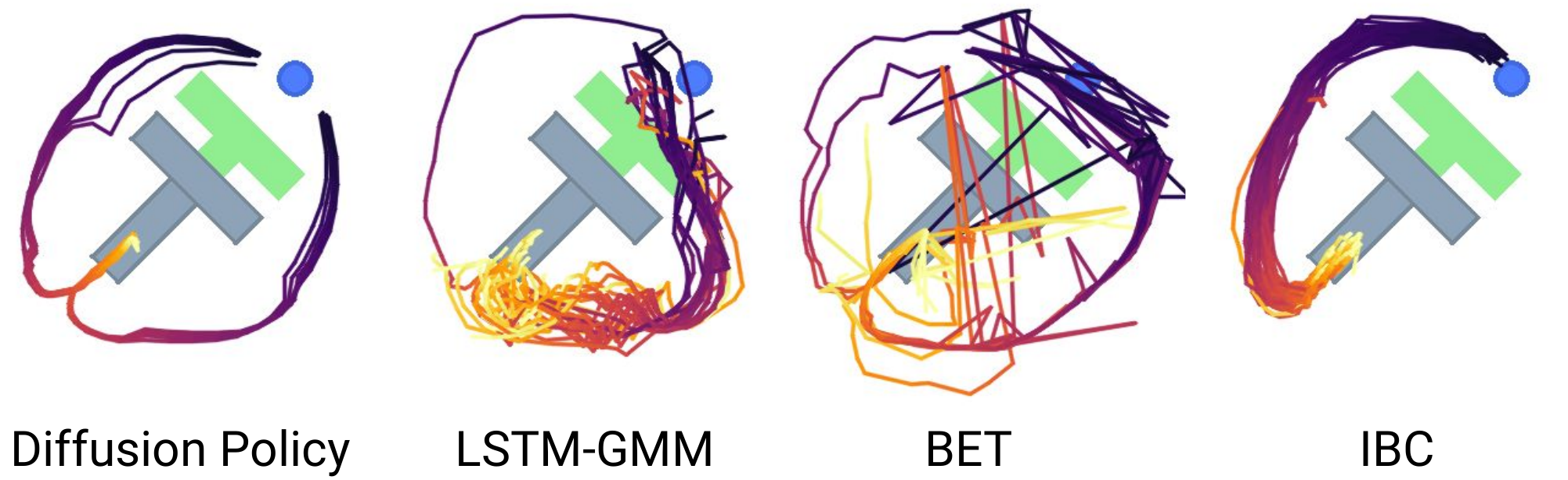} %\vspace{-3mm}

% https://docs.google.com/drawings/d/1Jmdy3Wga8EAxxDtma6kV0yvTbkoqADJy24FpFfCHk-4/edit
% \caption{
% \label{fig:multimodal}
% \textbf{Emergent multimodal behavior.} 
% The end-effector can go left or right to push the T block into the green target. 
% \textbf{Diffusion Policy} learned to detour around the block equally, 
% while \textbf{LSTM-GMM} generates actions heavily biased towards the right side. 
% \textbf{BET} generates jittery actions due to its lack of temporal action consistency and 
% \textbf{IBC} generates actions only from the left side due to mode collapse. 
% % These results were obtained by rolling out 40 steps for each method in Table 3.
% }
\caption{\label{fig:multimodal} 
\textbf{Multimodal behavior.} At the given state, the end-effector (blue) can either go left or right to push the block.
\textbf{Diffusion Policy} learns both modes and commits to only one mode within each rollout.  
% try to rehash same terms
In contrast, both \textbf{LSTM-GMM} \cite{robomimic} and \textbf{IBC} \cite{ibc} are biased toward one mode, while \textbf{BET} \cite{bet} fails to commit to a single mode due to its lack of temporal action consistency. 
Actions generated by rolling out 40 steps for the best-performing checkpoint. 
% \todo{Add diffusion process in the first row}
}
\vspace{-2mm}
\end{figure}

Intuitively, multi-modality in action generation for diffusion policy arises from two sources -- an underlying stochastic sampling procedure and a stochastic initialization. In Stochastic Langevin Dynamics, an initial sample $\mathbf{A}^K_t$ is drawn from standard Gaussian at the beginning of each sampling process, which helps specify different possible convergence basins for the final action prediction $\mathbf{A}^0_t$. This action is then further stochastically optimized, with added Gaussian perturbations across a large number of iterations, which enables individual action samples to converge and move between different multi-modal action basins.
Fig. \ref{fig:multimodal}, shows an example of the Diffusion Policy's multimodal behavior in a planar pushing task (Push T, introduced below) without explicit demonstration for the tested scenario. 

\subsection{Synergy with Position Control} 
\label{sec:property_pos_vs_vel}
\begin{figure}[h]
\centering
\includegraphics[width=0.85\linewidth]{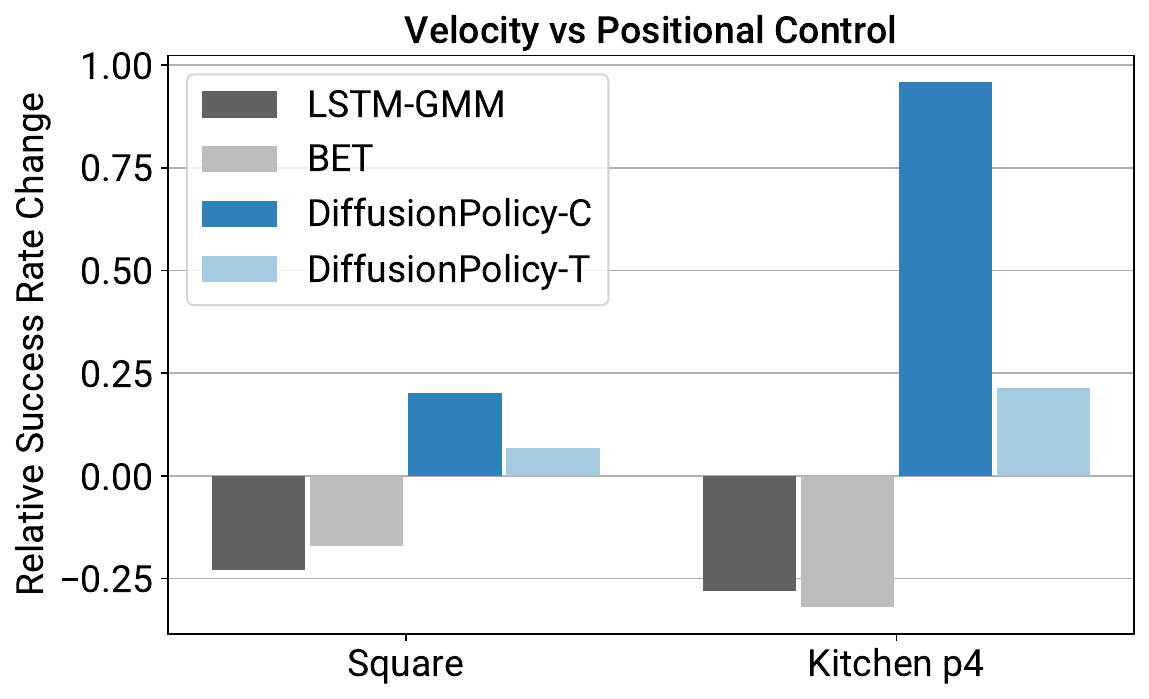}
\caption{\textbf{Velocity v.s. Position Control.} \label{fig:pos_vs_vel} The performance difference when switching from velocity to position control. While both BCRNN and BET performance decrease, Diffusion Policy is able to leverage the advantage of position and improve its performance. }
\vspace{-4mm}
\end{figure}

We find that Diffusion Policy with a position-control action space consistently outperforms Diffusion Policy with velocity control, as shown in Fig \ref{fig:pos_vs_vel}. This surprising result stands in contrast to the majority of recent behavior cloning work that generally relies on velocity control \cite{robomimic, bet, zhang2018deep, florence2019self, mandlekar2020learning, mandlekar2020iris}. We speculate that there are two primary reasons for this discrepancy: First, action multimodality is more pronounced in position-control mode than it is when using velocity control. Because Diffusion Policy better expresses action multimodality than existing approaches, we speculate that it is inherently less affected by this drawback than existing methods. Furthermore, position control suffers less than velocity control from compounding error effects and is thus more suitable for action-sequence prediction (as discussed in the following section). As a result, Diffusion Policy is both less affected by the primary drawbacks of position control and is better able to exploit position control's advantages.

\subsection{Benefits of Action-Sequence Prediction}
\label{sec:action_sequence}

%Sec. \ref{sec:multimodal} discusses how Diffusion Policy is able to capture multimodal distribution in a single action step. However, it is not sufficient for modeling the temporary consistency between the action step. 

Sequence prediction is often avoided in most policy learning methods due to the difficulties in effectively sampling from high-dimensional output spaces. For example, IBC would struggle in effectively sampling high-dimensional action space with a non-smooth energy landscape. Similarly, BC-RNN and BET would have difficulty specifying the number of modes that exist in the action distribution (needed for GMM or k-means steps).  

In contrast, DDPM scales well with output dimensions without sacrificing the expressiveness of the model, as demonstrated in many image generation applications. Leveraging this capability, Diffusion Policy represents action in the form of a high-dimensional action sequence, which naturally addresses the following issues: 

\begin{itemize} [leftmargin=3mm]
    \item \textbf{Temporal action consistency}: Take Fig \ref{fig:multimodal} as an example. To push the T block into the target from the bottom, the policy can go around the T block from either left or right. However, suppose each action in the sequence is predicted as independent multimodal distributions (as done in BC-RNN and BET). In that case, consecutive actions could be drawn from different modes, resulting in jittery actions that alternate between the two valid trajectories. 

    \item \textbf{Robustness to idle actions}: Idle actions occur when a demonstration is paused and results in sequences of identical positional actions or near-zero velocity actions. It is common during teleoperation and is sometimes required for tasks like liquid pouring. However, single-step policies can easily overfit to this pausing behavior. For example, BC-RNN and IBC often get stuck in real-world experiments when the idle actions are not explicitly removed from training. 
    
\end{itemize}

\begin{figure}[h]
\centering
\includegraphics[width=\linewidth]{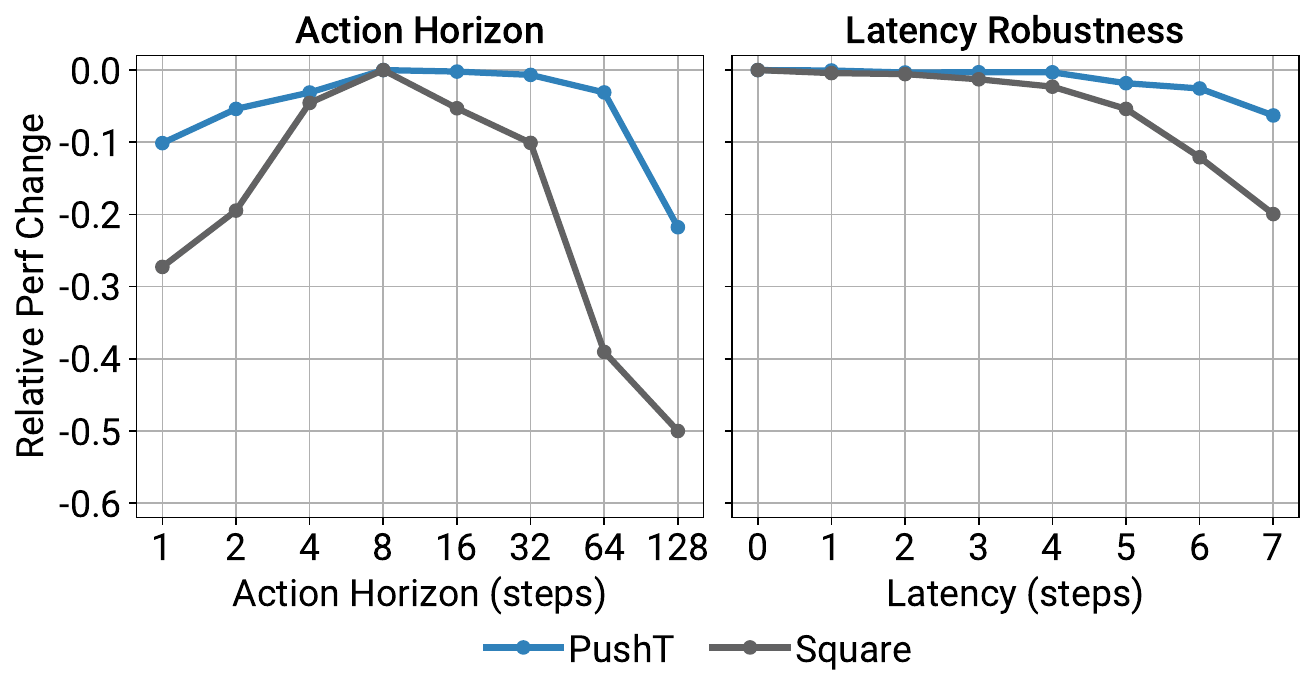}
\vspace{-6mm}

\caption{\textbf{Diffusion Policy Ablation Study.} 
Change (difference) in success rate relative to the maximum for each task is shown on the Y-axis.
\textbf{Left}: trade-off between temporal consistency and responsiveness when selecting the action horizon. 
\textbf{Right}: Diffusion Policy with position control is robust against latency.
Latency is defined as the number of steps between the last frame of observations to the first action that can be executed.
} 
% (b) The impact of observation horizon length on task performance.(d) Diffusion Policy is more sample efficient than BCRNN.
\label{fig:ablation}
\vspace{-5mm}
\end{figure}

\subsection{Training Stability}
\label{sec:ibc_stability}
%IBC \cite{ibc} has demonstrated impressive performance on tasks reported in the paper. However, alongside other works like \cite{bet}, we struggled to achieve the same level of performance on some other tasks, such as the robomimc suite and block pushing.  We believe the performance disparity can be explained by IBC's instability training stability. %When training on the same simulated Push T task (best performing task for IBC), the evaluation success rate for IBC oscillates in a wide range throughout the training process. This makes selecting a good IBC checkpoint for realworld tasks very difficult.

\begin{figure}[h]
\centering
% \vspace{-2mm}
\includegraphics[width=\linewidth]{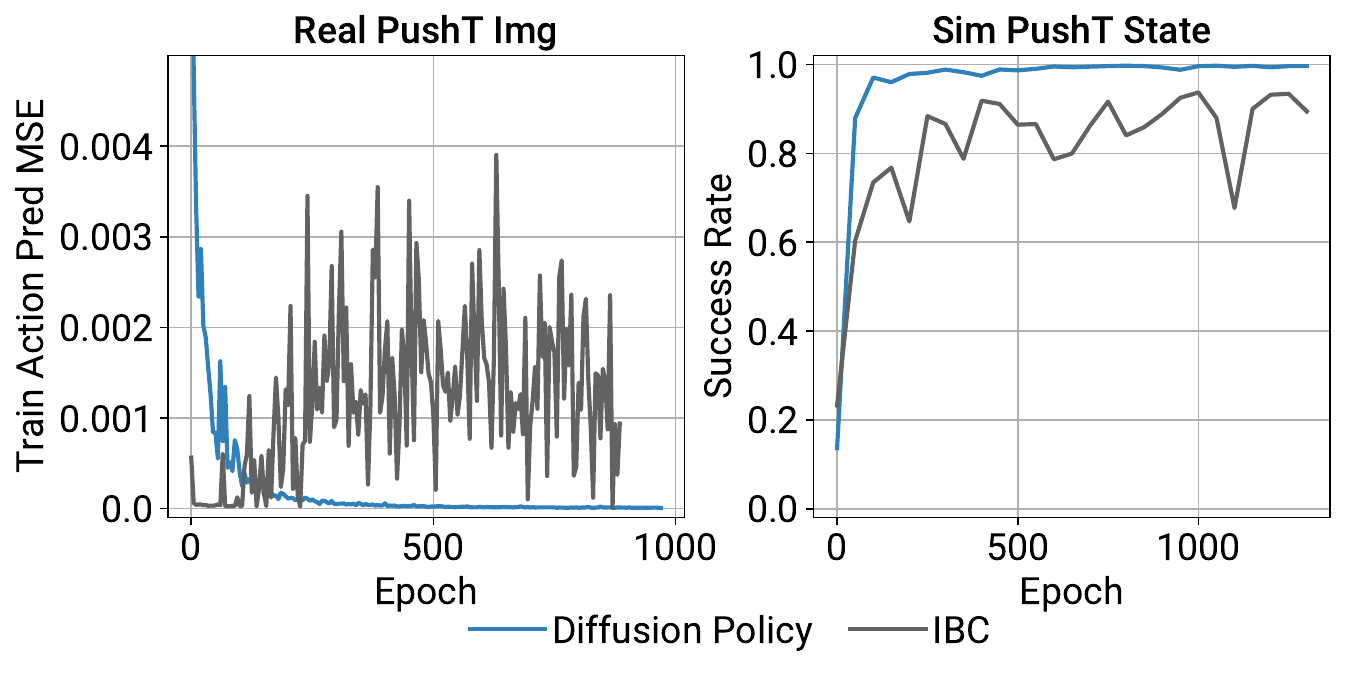}
\vspace{-5mm}
\caption{\textbf{Training Stability.} \label{fig:ibc_stability}  Left: IBC fails to infer training actions with increasing accuracy despite smoothly decreasing training loss for energy function. Right: IBC's evaluation success rate oscillates, making checkpoint selection difficult (evaluated using policy rollouts in simulation).}

%\vspace{-9.5mm}
\end{figure}

While IBC, in theory, should possess similar advantages as diffusion policies. However, achieving reliable and high-performance results from IBC in practice is challenging due to IBC's inherent training instability \cite{ta2022conditional}. Fig \ref{fig:ibc_stability} shows training error spikes and unstable evaluation performance throughout the training process, making hyperparameter turning critical and checkpoint selection difficult. As a result, \citet{ibc} evaluate every checkpoint and report results for the best-performing checkpoint. In a real-world setting, this workflow necessitates the evaluation of many policies on hardware to select a final policy. Here, we discuss why Diffusion Policy appears significantly more stable to train.

An implicit policy represents the action distribution using an Energy-Based Model (EBM):
\vspace{-2mm}
\begin{equation}
    \label{eq:ebm}
    p_\theta(\mathbf{a}|\mathbf{o})=\frac{e^{-E_\theta(\mathbf{o},\mathbf{a})}}{Z(\mathbf{o},\theta)}
\vspace{-1mm}
\end{equation}
where $Z(\mathbf{o},\theta)$ is an intractable normalization constant (with respect to $\mathbf{a}$). 

To train the EBM for implicit policy, an InfoNCE-style loss function is used, which equates to the negative log-likelihood of Eq \ref{eq:ebm}:
% \vspace{-3mm}
\begin{equation}
    \label{eq:infonce}
    \mathcal{L}_{infoNCE}=-\log(\frac{
        e^{-E_\theta(\mathbf{o},\mathbf{a})}
    }{
        e^{-E_\theta(\mathbf{o},\mathbf{a})} + 
            \textcolor{red}{\sum^{N_{neg}}_{j=1}}e^{
                -E_\theta(\mathbf{o},\textcolor{red}{\widetilde{\mathbf{a}}^j})}
    })
\end{equation}
where a set of negative samples $\textcolor{red}{\{\widetilde{\mathbf{a}}^j\}^{N_{neg}}_{j=1}}$ are used to estimate the intractable normalization constant $Z(\mathbf{o},\theta)$. In practice, the inaccuracy of negative sampling is known to cause training instability for EBMs \cite{du2020improved,ta2022conditional}.

Diffusion Policy and DDPMs sidestep the issue of estimating $Z(\mathbf{a},\theta)$ altogether by modeling the \textbf{score function} \cite{song2019score} of the same action distribution in Eq \ref{eq:ebm}:
\begin{equation}
    \nabla_{\mathbf{a}}\log p(\mathbf{a}|\mathbf{o})
    =-\nabla_{\mathbf{a}} E_{\theta}(\mathbf{a},\mathbf{o})-\underbrace{\nabla_{\mathbf{a}}\log Z(\mathbf{o},\theta)}_{=0}
    % =-\nabla_{\mathbf{a}} E_{\theta}(\mathbf{a},\mathbf{o})
    \approx -\epsilon_\theta(\mathbf{a},\mathbf{o})
\end{equation}
where the noise-prediction network $\epsilon_\theta(\mathbf{a},\mathbf{o})$ is approximating the negative of the score function $\nabla_{\mathbf{a}}\log p(\mathbf{a}|\mathbf{o})$ \cite{liu2022compositional}, which is independent of the normalization constant $Z(\mathbf{o},\theta)$. As a result, neither the inference (Eq \ref{eq:diffusion_policy_langevin}) nor training (Eq \ref{eq:diffusion_policy_loss}) process of Diffusion Policy involves evaluating $Z(\mathbf{o},\theta)$, thus making Diffusion Policy training more stable.

% An implicit policy represents the probability distribution across different actions $\mathbf{a}$ using an EBM
% \begin{equation}
%     p_\theta(\mathbf{a}) \propto e^{-E_\theta(\mathbf{a})}.
% \end{equation}
% where actions are generated by running MCMC on $p_\theta(\mathbf{a})$. One common approach to sample from EBMs is Langevin Dynamics:  
% \begin{equation}
%     \label{eq:langevin}
%     \mathbf{a}_k = 
%     \mathbf{a}_{t - 1} - 
%     \nabla_{\mathbf{a}} E_{\theta}(\mathbf{a}_{k-1},\mathbf{o}) +
%     \mathcal{N}(0, \sigma_k^2 I),
% \end{equation}
% where the $\nabla_{\mathbf{a}} E_{\theta}(\mathbf{a}_{k-1},\mathbf{o})$ is used to gradually denoise actions. Equation \ref{eq:langevin} is functionally quite similar to \ref{eq:diffusion_policy_langevin}, with $\nabla_{\mathbf{a}} E_{\theta}(\mathbf{a}_{k-1})$
% replacing the noise prediction network $\epsilon_\theta(\mathbf{a}_k, k)$.

% Indeed as discussed in \citet{liu2022compositional}, the perturbation function  $\epsilon_\theta(\mathbf{a}_k, k)$ directly learns to estimate the gradient field $\nabla_{\mathbf{a}} E_{\theta}(\mathbf{a}_{k-1})$. Thus a DDPM model is further an EBM, and directly parameterizes an implicit policy model through a set of learned gradient fields representing how individual actions should be refined. In comparison to existing approaches to training implicit policies, which rely on NCE loss \cite{ibc}, this DDPM objective may be seen as a more stable analogue to train such an implicit policy, where gradient fields are densely supervised through denoising.

\subsection{Connections to Control Theory}
\label{sec:control}
Diffusion Policy has a simple limiting behavior when the tasks are very simple; this potentially allows us to bring to bear some rigorous understanding from control theory. Consider the case where we have a linear dynamical system, in standard state-space form, that we wish to control:
\begin{gather*} 
{\bf s}_{t+1} = {\bf A}{\bf s}_t + {\bf B}{\bf a}_t + {\bf w}_t, \qquad {\bf w}_t \sim \mathcal{N}(0, \Sigma_w).
%, \\ {\bf o}_t = {\bf C}{\bf s}_t + {\bf v}_t, \qquad {\bf v}_t \sim \mathcal{N}(0, \Sigma_v).
\end{gather*} Now imagine we obtain demonstrations (rollouts) from a linear feedback policy: ${\bf a}_t = -{\bf K}{\bf s}_t.$ This policy could be obtained, for instance, by solving a linear optimal control problem like the Linear Quadratic Regulator. Imitating this policy does not need the modeling power of diffusion, but as a sanity check, we can see that Diffusion Policy does the right thing.

In particular, when the prediction horizon is one time step, $T_p=1$, it can be seen that the optimal denoiser which minimizes 
\begin{equation}
    \mathcal{L}=MSE(\mathbf{\epsilon}^k,\epsilon_\theta(\mathbf{s}_t, -{\bf K}{\bf s}_t + \mathbf{\epsilon}^k, k))
\end{equation}
is given by $$\epsilon_\theta({\bf s}, {\bf a}, k) = \frac{1}{\sigma_k}[{\bf a} + {\bf K}{\bf s}],$$ where $\sigma_k$ is the variance on denoising iteration $k$. Furthermore, at inference time, the DDIM sampling will converge to the global minima at ${\bf a} = -{\bf Ks}.$

Trajectory prediction ($T_p>1$) follows naturally. In order to predict ${\bf a}_{t+t'}$ as a function of ${\bf s}_t$, the optimal denoiser will produce ${\bf a}_{t+t'} = -{\bf K}({\bf A}-{\bf BK})^{t'}{\bf s}_t$; all terms involving ${\bf w}_t$ are zero in expectation. This shows that in order to perfectly clone a behavior that depends on the state, the learner must implicitly learn a (task-relevant) dynamics model \cite{subramanian2019approximate,zhang2020learning}. Note that if either the plant or the policy is nonlinear, then predicting future actions could become significantly more challenging and once again involve multimodal predictions.

%In the case of output feedback, one might like to understand if Diffusion Policy can recover e.g. Linear Quadratic Gaussian (LQG) optimal control. The Diffusion Policy architecture presented here takes only a finite history of observations, while the Kalman filter component of the LQG controller has an internal state. However, Diffusion Policy can recover a \emph{truncated} version of the unrolled LQG controller, as could be obtained from e.g. System-Level Synthesis \cite{anderson2019system}. In general, this truncated LQG should be given access to the history of \emph{actions} in addition to observations in order to more closely approximate the optimal LQG (the Kalman filter state is a function of both previous actions and observations). In this setting it may be helpful to think of the intermediate representations learned by Diffusion Policy as a compressed, task-relevant belief state.
%$$a_t = \sum_{i=0}^{T_o} K_i y_{t-i} + \sum_{i=1}^{T_o} \bar{K}_i a_{t-1}.$$
\begin{table*}[h]
    
~~~~~~~~~~~~~~~~~~~~~~~~~~~~~~~~~~~~
\includegraphics[width=0.835\linewidth]{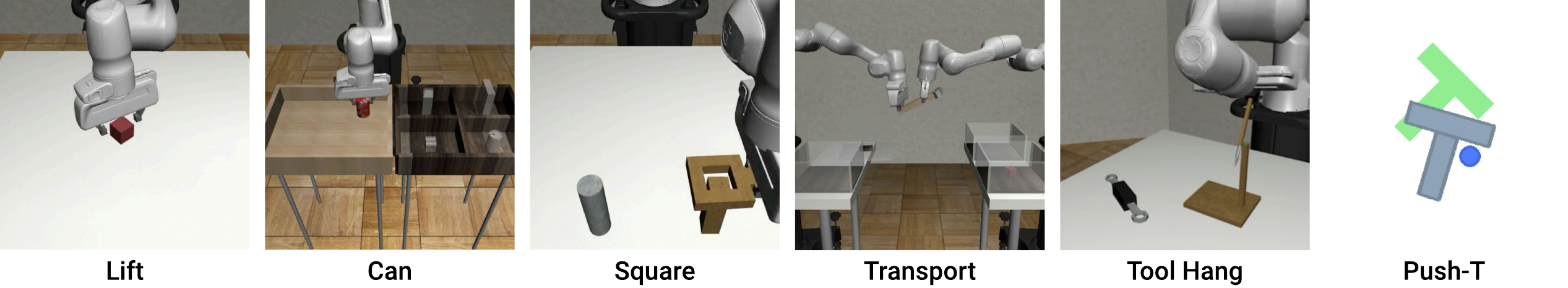}
\label{tab:sim_benchmark_state}
% https://docs.google.com/drawings/d/11Azs8uUlZ2CER5NarlBOTnGhj3o_jm59wIqQKW3rXNQ/edit

\vspace{1mm}
{
\centering
\setlength\tabcolsep{ 3 pt}
\begin{tabular}{r|cc|cc|cc|cc|c|c}
\toprule
 & \multicolumn{2}{c|}{Lift} & \multicolumn{2}{c|}{Can} & \multicolumn{2}{c|}{Square} & \multicolumn{2}{c|}{Transport} & \multicolumn{1}{c|}{ToolHang} & \multicolumn{1}{c}{Push-T} \\
 & ph & mh & ph & mh & ph & mh & ph & mh & ph & ph \\
\midrule
LSTM-GMM & \small \textbf{1.00}/0.96 & \small \textbf{1.00}/0.93 & \small \textbf{1.00}/0.91 & \small \textbf{1.00}/0.81 & \small 0.95/0.73 & \small 0.86/0.59 & \small 0.76/0.47 & \small 0.62/0.20 & \small 0.67/0.31 & \small 0.67/0.61 \\
IBC & \small 0.79/0.41 & \small 0.15/0.02 & \small 0.00/0.00 & \small 0.01/0.01 & \small 0.00/0.00 & \small 0.00/0.00 & \small 0.00/0.00 & \small 0.00/0.00 & \small 0.00/0.00 & \small 0.90/0.84 \\
BET & \small \textbf{1.00}/0.96 & \small \textbf{1.00}/0.99 & \small \textbf{1.00}/0.89 & \small \textbf{1.00}/0.90 & \small 0.76/0.52 & \small 0.68/0.43 & \small 0.38/0.14 & \small 0.21/0.06 & \small 0.58/0.20 & \small 0.79/0.70 \\
\midrule
DiffusionPolicy-C & \small \textbf{1.00}/0.98 & \small \textbf{1.00}/0.97 & \small \textbf{1.00}/0.96 & \small \textbf{1.00}/\textbf{0.96} & \small \textbf{1.00}/\textbf{0.93} & \small \textbf{0.97}/\textbf{0.82} & \small 0.94/0.82 & \small \textbf{0.68}/\textbf{0.46} & \small 0.50/0.30 & \small 0.95/\textbf{0.91} \\
DiffusionPolicy-T & \small \textbf{1.00}/\textbf{1.00} & \small \textbf{1.00}/\textbf{1.00} & \small \textbf{1.00}/\textbf{1.00} & \small \textbf{1.00}/0.94 & \small \textbf{1.00}/0.89 & \small 0.95/0.81 & \small \textbf{1.00}/\textbf{0.84} & \small 0.62/0.35 & \small \textbf{1.00}/\textbf{0.87} & \small \textbf{0.95}/0.79 \\
\bottomrule
\end{tabular}

\caption{\textbf{Behavior Cloning Benchmark (State Policy) \label{tab:table_low_dim} } 
We present success rates with different checkpoint selection methods in the format of (max performance) / (average of last 10 checkpoints), with each averaged across 3 training seeds and 50 different environment initial conditions (150 in total). 
LSTM-GMM corresponds to BC-RNN in RoboMimic\cite{robomimic}, which we reproduced and obtained slightly {better} results than the original paper. Our results show that Diffusion Policy significantly improves state-of-the-art performance across the board.
% \label{tab:table_low_dim}
% Numbers are success rate shown with different checkpoint selection methods: (max performance) / (average of last 10 checkpoints), which are then averaged across 3 training seeds and 50 different env initial conditions each (150 in total). Higher the better. LSTM-GMM correspond to BC-RNN in RoboMimic \cite{robomimic}. LSTM-GMM numbers are from our reproduction (generally slightly higher than the original paper).  \label{tab:table_low_dim}
}
\vspace{2mm}

\setlength\tabcolsep{ 3 pt}
\begin{tabular}{r|cc|cc|cc|cc|c|c}
\toprule
 & \multicolumn{2}{c|}{Lift} & \multicolumn{2}{c|}{Can} & \multicolumn{2}{c|}{Square} & \multicolumn{2}{c|}{Transport} & \multicolumn{1}{c|}{ToolHang} & \multicolumn{1}{c}{Push-T} \\
 & ph & mh & ph & mh & ph & mh & ph & mh & ph & ph \\
\midrule
LSTM-GMM & \small \textbf{1.00}/0.96 & \small \textbf{1.00}/0.95 & \small \textbf{1.00}/0.88 & \small 0.98/0.90 & \small 0.82/0.59 & \small 0.64/0.38 & \small 0.88/0.62 & \small 0.44/0.24 & \small 0.68/0.49 & \small 0.69/0.54 \\
IBC & \small 0.94/0.73 & \small 0.39/0.05 & \small 0.08/0.01 & \small 0.00/0.00 & \small 0.03/0.00 & \small 0.00/0.00 & \small 0.00/0.00 & \small 0.00/0.00 & \small 0.00/0.00 & \small 0.75/0.64 \\
DiffusionPolicy-C & \small \textbf{1.00}/\textbf{1.00} & \small \textbf{1.00}/\textbf{1.00} & \small \textbf{1.00}/0.97 & \small \textbf{1.00}/0.96 & \small 0.98/\textbf{0.92} & \small \textbf{0.98}/\textbf{0.84} & \small \textbf{1.00}/\textbf{0.93} & \small \textbf{0.89}/\textbf{0.69} & \small \textbf{0.95}/\textbf{0.73} & \small \textbf{0.91}/\textbf{0.84} \\
DiffusionPolicy-T & \small \textbf{1.00}/\textbf{1.00} & \small \textbf{1.00}/0.99 & \small \textbf{1.00}/\textbf{0.98} & \small \textbf{1.00}/\textbf{0.98} & \small \textbf{1.00}/0.90 & \small 0.94/0.80 & \small 0.98/0.81 & \small 0.73/0.50 & \small 0.76/0.47 & \small 0.78/0.66 \\
\bottomrule
\end{tabular}

\caption{\textbf{Behavior Cloning Benchmark (Visual Policy) \label{tab:table_image}} Performance are reported in the same format as in Tab \ref{tab:table_low_dim}. LSTM-GMM numbers were reproduced to get a complete evaluation in addition to the best checkpoint performance reported. Diffusion Policy shows consistent performance improvement, especially for complex tasks like Transport and ToolHang. }
}
%\vspace{-3mm}
\end{table*}

\section{Evaluation}
We systematically evaluate Diffusion Policy on 15 tasks from 4 benchmarks \cite{ibc, gupta2019relay, robomimic, bet}. This evaluation suite includes both simulated and real environments, single and multiple task benchmarks, fully actuated and under-actuated systems, and rigid and fluid objects.  We found Diffusion Policy to consistently outperform the prior state-of-the-art on all of the tested benchmarks, with an average success-rate improvement of 46.9\%. In the following sections, we  provide an overview of each task, our evaluation methodology on that task, and our key takeaways.

\subsection{Simulation Environments and datasets}
\textbf{Robomimic}
\cite{robomimic} is a large-scale robotic manipulation benchmark designed to study imitation learning and offline RL. The benchmark consists of 5 tasks with a proficient human (PH) teleoperated demonstration dataset for each and mixed proficient/non-proficient human (MH) demonstration datasets for 4 of the tasks (9 variants in total). For each variant, we report results for both state- and image-based observations. Properties for each task are summarized in Tab \ref{tab:robomimic_tasks}.

\begin{table}
\centering
\setlength\tabcolsep{2 pt}
\small
\begin{tabular}{c|cccccccc}
\toprule
Task      & \# Rob & \# Obj & ActD & \#PH & \#MH & Steps & Img? & HiPrec \\
\midrule
\multicolumn{1}{c}{} & \multicolumn{8}{c}{Simulation Benchmark} \\
% \midrule
\midrule
Lift      & 1      & 1      & 7    & 200   & 300   & 400       & Yes    & No       \\
Can       & 1      & 1      & 7    & 200   & 300   & 400       & Yes    & No       \\
Square    & 1      & 1      & 7    & 200   & 300   & 400       & Yes    & Yes      \\
Transport & 2      & 3      & 14   & 200   & 300   & 700       & Yes    & No       \\
ToolHang  & 1      & 2      & 7    & 200   & 0     & 700       & Yes    & Yes      \\
\midrule
Push-T    & 1      & 1      & 2    & 200   & 0     & 300       & Yes    & Yes      \\
\midrule
BlockPush & 1      & 2      & 2    & 0     & 0     & 350       & No     & No       \\
\midrule
Kitchen   & 1      & 7      & 9    & 656   & 0     & 280       & No     & No      \\
\midrule
\multicolumn{1}{c}{} & \multicolumn{8}{c}{Realworld Benchmark} \\
\midrule
Push-T    & 1      & 1      & 2    & 136   & 0     & 600       & Yes    & Yes  \\
6DoF Pour   & 1    & liquid      & 6    & 90   & 0     & 600       & Yes    & No  \\
Peri Spread   & 1      & liquid      & 6    & 90   & 0     & 600       & Yes    & No  \\
Mug Flip  & 1      & 1      & 7    & 250   & 0     & 600       & Yes    & No  \\
\bottomrule
\end{tabular}

\caption{\textbf{Tasks Summary.} \# Rob: number of robots, \#Obj: number of objects, ActD: action dimension, PH: proficient-human demonstration, MH: multi-human demonstration, Steps: max number of rollout steps, HiPrec: whether the task has a high precision requirement. BlockPush uses 1000 episodes of scripted  demonstrations.}
%The 5 robomimic \cite{robomimic} tasks and the Push-T have both state-based and vision-based tasks. 4 of the 5 robomimic tasks has both proficient-human ``PH'' and multi-human ``MH'' datasets. The Block Pushing task has 1000 machine-generated demonstrations instead. The Push-T environment allows an eval episode to be terminated earlier than 300 steps.
\label{tab:robomimic_tasks} 
\vspace{-5mm}
\end{table}

\textbf{Push-T}
\label{sec:eval_sim_pusht}
adapted from IBC \cite{ibc}, requires pushing a T-shaped block (gray) to a fixed target (red) with a circular end-effector (blue)s. Variation is added by random initial conditions for T block and end-effector. The task requires exploiting complex and contact-rich object dynamics to push the T block precisely, using point contacts. There are two variants: one with RGB image observations and another with 9 2D keypoints obtained from the ground-truth pose of the T block, both with proprioception for end-effector location.
% is a task we adopted from IBC \cite{ibc} to explore how Diffusion Policy work under the complex, contact-rich, and underactuated dynamics. The task is to push a T-shaped block (colored gray) into a fixed target pose (colored red) with a fixed circular end-effector (colored blue). The variation between episodes comes from the random initial condition for both the T block and the end-effector. Since the T block can only be pushed around by point contacts, the agent needs to exploit the object dynamics and constantly to change contact points and strategies to push the T block into a precise location. The task as two variants, one uses RGB image-based observation, and another is state-based, where 9 fixed 2D keypoints are computed from the ground truth pose for the T block. In addition, each variant also provides the current end-effector location for proprioception.

\textbf{Multimodal Block Pushing} adapted from BET \cite{bet}, this task tests the policy's ability to model multimodal action distributions by pushing two blocks into two squares in any order. The demonstration data is generated by a scripted oracle with access to groundtruth state info. This oracle randomly selects an initial block to push and moves it to a randomly selected square. The remaining block is then pushed into the remaining square. This task contains \textbf{long-horizon} multimodality that can not be modeled by a single function mapping from observation to action.

\textbf{Franka Kitchen} is a popular environment for evaluating the ability of IL and Offline-RL methods to learn multiple long-horizon tasks. Proposed in Relay Policy Learning \cite{gupta2019relay}, the Franka Kitchen environment contains 7 objects for interaction and comes with a human demonstration dataset of 566 demonstrations, each completing 4 tasks in arbitrary order. The goal is to execute as many demonstrated tasks as possible, regardless of order, showcasing both short-horizon and long-horizon multimodality.

\begin{table}[t]
\centering
\includegraphics[width=0.9\linewidth]{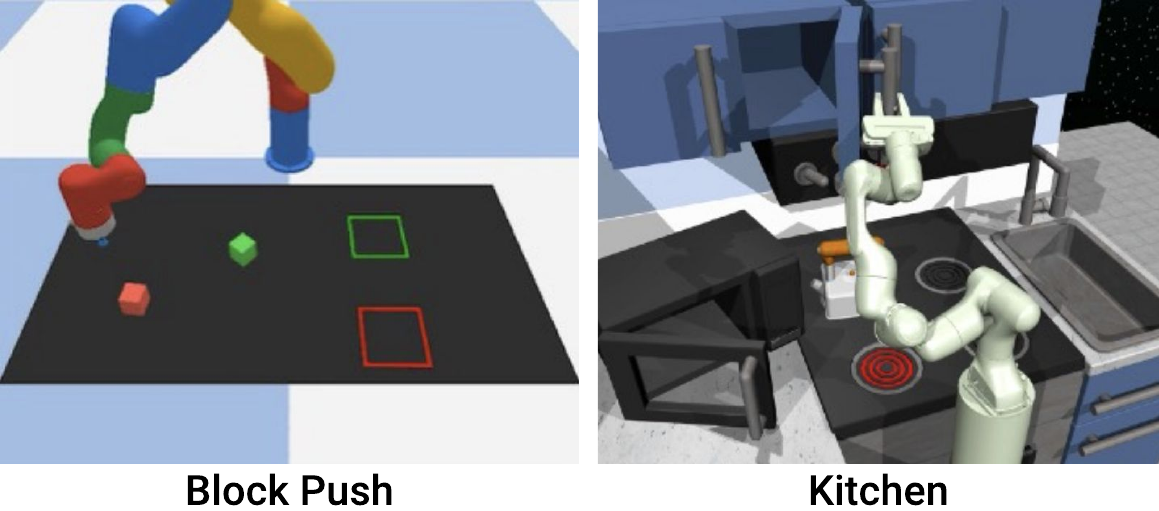}

\vspace{2mm}
\setlength\tabcolsep{4.8 pt}
\begin{tabular}{r|cc|cccc}
\toprule
 & \multicolumn{2}{c|}{BlockPush} & \multicolumn{4}{c}{Kitchen} \\
 & p1 & p2 & p1 & p2 & p3 & p4 \\
\midrule
LSTM-GMM & \small 0.03 & \small 0.01 & \small \textbf{1.00} & \small 0.90 & \small 0.74 & \small 0.34 \\
IBC & \small 0.01 & \small 0.00 & \small 0.99 & \small 0.87 & \small 0.61 & \small 0.24 \\
BET & \small 0.96 & \small 0.71 & \small 0.99 & \small 0.93 & \small 0.71 & \small 0.44 \\
DiffusionPolicy-C & \small 0.36 & \small 0.11 & \small \textbf{1.00} & \small \textbf{1.00} & \small \textbf{1.00} & \small \textbf{0.99} \\
DiffusionPolicy-T & \small \textbf{0.99} & \small \textbf{0.94} & \small \textbf{1.00} & \small 0.99 & \small 0.99 & \small 0.96 \\
\bottomrule
\end{tabular}

\caption{\textbf{Multi-Stage Tasks (State Observation)}. 
\label{tab:multi_stage}
For PushBlock, $px$ is the frequency of pushing $x$ blocks into the targets. 
For Kitchen, $px$ is the frequency of interacting with $x$ or more objects (e.g. bottom burner). 
Diffusion Policy performs better, especially for difficult metrics such as $p2$ for Block Pushing and $p4$ for Kitchen, as demonstrated by our results.
}
\vspace{-4mm}
\end{table}

\subsection{Evaluation Methodology}
We present the \textbf{best-performing for each baseline method} on each benchmark from all possible sources -- our reproduced result (LSTM-GMM) or original number reported in the paper (BET, IBC). We report results from the average of the last 10 checkpoints (saved every 50 epochs) across \textbf{3} training seeds and \textbf{50} environment initializations 
\footnote{Due to a bug in our evaluation code, only 22 environment initializations are used for robomimic tasks. This does not change our conclusion since all baseline methods are evaluated in the same way.} 
(an average of \textbf{1500} experiments in total). The metric for most tasks is success rate, except for the Push-T task, which uses target area coverage.
In addition, we report the average of best-performing checkpoints for robomimic and Push-T tasks to be consistent with the evaluation methodology of their respective original papers \cite{robomimic, ibc}. All state-based tasks are trained for 4500 epochs, and image-based tasks for 3000 epochs. Each method is evaluated with its best-performing action space: position control for Diffusion Policy and velocity control for baselines (the effect of action space will be discussed in detail in Sec \ref{sec:eval_pos_vs_vel}).  
The results from these simulation benchmarks are summarized in Table \ref{tab:table_low_dim} and Table \ref{tab:table_image}.

\subsection{Key Findings}

Diffusion Policy outperforms alternative methods on all tasks and variants, with both state and vision observations, in our simulation benchmark study (Tabs \ref{tab:table_low_dim}, \ref{tab:table_image} and \ref{tab:multi_stage}) with an average improvement of 46.9\%. The following paragraphs summarize the key takeaways.

\textbf{Diffusion Policy can express short-horizon multimodality.}
We define short-horizon action multimodality as multiple ways of achieving \textbf{the same immediate goal}, which is prevalent in human demonstration data \cite{robomimic}. %For example, when navigating around a fixed obstacle, one may choose to traverse it on the left or the right leading to short-horizon action multimodality.
In Fig \ref{fig:multimodal}, we present a case study of this type of short-horizon multimodality in the Push-T task. Diffusion Policy learns to approach the contact point equally likely from left or right, while LSTM-GMM \cite{robomimic} and IBC \cite{ibc} exhibit bias toward one side and BET \cite{bet} cannot commit to one mode.

\textbf{Diffusion Policy can express long-horizon multimodality.}
Long-horizon multimodality is the completion of \textbf{different sub-goals} in inconsistent order. For example, the order of pushing a particular block in the Block Push task or the order of interacting with 7 possible objects in the Kitchen task are arbitrary.
We find that Diffusion Policy copes well with this type of multimodality; it outperforms baselines on both tasks by a large margin: 32\% improvement on Block Push's p2 metric and 213\% improvement on Kitchen's p4 metric.

\textbf{Diffusion Policy can better leverage position control.}
\label{sec:eval_pos_vs_vel}
Our ablation study (Fig. \ref{fig:pos_vs_vel}) shows that selecting position control as the diffusion-policy action space significantly outperformed velocity control. The baseline methods we evaluate, however, work best with velocity control (and this is reflected in the literature where most existing work reports using velocity-control action spaces \cite{robomimic, bet, zhang2018deep, florence2019self, mandlekar2020learning, mandlekar2020iris}).

%However, position control action space hurts performance for baseline methods, which explains why most existing work on behavior cloning for manipulation use velocity control \cite{robomimic, bet, zhang2018deep, florence2019self, mandlekar2020learning, mandlekar2020iris}.

% Most existing work on behavior cloning for robotic manipulation uses some variant of velocity control (e.g. delta pose relative to the end-effector pose in the previous time step) as action space \cite{robomimic, bet, zhang2018deep, florence2019self, mandlekar2020learning, mandlekar2020iris}. Our ablation study as well as our experience in tuning baseline methods confirm that velocity control is the better action space for baseline methods, with performance penalty of switching to position action space at around 20\%. In contrast, we found switching to position control leads to significant performance improvement for Diffusion Policy as shown in Fig. \ref{fig:pos_vs_vel}. The choice of action space is discussed in detail in Sec \ref{sec:property_pos_vs_vel}.

\textbf{The tradeoff in action horizon.}
As discussed in Sec \ref{sec:action_sequence}, 
having an action horizon greater than 1 helps the policy predict consistent actions and compensate for idle portions of the demonstration, but too long a horizon reduces performance due to slow reaction time. Our experiment confirms this trade-off (Fig. \ref{fig:ablation} left) and found the action horizon of 8 steps to be optimal for most tasks that we tested. 

\textbf{Robustness against latency.}
Diffusion Policy employs receding horizon position control to predict a sequence of actions into the future. This design helps address the latency gap caused by image processing, policy inference, and network delay. Our ablation study with simulated latency showed Diffusion Policy is able to maintain peak performance with latency up to 4 steps (Fig \ref{fig:ablation}). We also find that velocity control is more affected by latency than position control, likely due to compounding error effects.

% Diffusion Policy's receding horizon control formulation allows predicting more actions than what's necessary for immediate execution. These extra actions predicted by previous policy execution can fill in the gap between the start of the current policy inference and when the first predicted action can be executed by the robot, which enables Diffusion Policy to be robust against latency. 

\textbf{Diffusion Policy is stable to train.}
We found that the optimal hyperparameters for Diffusion Policy are mostly consistent across tasks.  In contrast,  IBC \cite{ibc} is prone to training instability. This property is discussed in Sec \ref{sec:ibc_stability}.

% move to appendix
% \begin{figure}[h]
% \centering
% \includegraphics[width=\linewidth]{figure/sample_efficiency_figure.pdf}

% \caption{\textbf{Data Efficiency.} 
% \label{fig:data_efficiency}
% Percentage change in success rate relative to the maximum (for both methods) is shown on the Y-axis.
% % \textbf{Left}: trade-off between temporal consistency and responsiveness when selecting the action horizon. 
% % \textbf{Right}: Diffusion Policy with position control is more robust against latency than velocity control.
% % (b) The impact of observation horizon length on task performance.(d) Diffusion Policy is more sample efficient than BCRNN.
% }
% \end{figure}

\subsection{Ablation Study}
\label{sec:arch_ablation}
We explore alternative vision encoder design decisions on the simulated robomimic square task.
Specifically, we evaluated 3 different architectures: 
ResNet-18, ResNet-34 \cite{resnet} 
and ViT-B/16 \cite{dosovitskiy2020image}. 
For each architecture, we evaluated 3 different training strategies:
training end-to-end from scratch,
using frozen pre-trained vision encoder, 
and finetuning pre-trained vision encoders (with 10x lower learning rate with respect to the policy network).
We use ImageNet-21k \cite{ridnik2021imagenet21k} pretraining for ResNet and CLIP \cite{radford2021learning} pretraining for ViT-B/16.
The quantitative comparison on square task with proficient-human (PH) dataset is shown in Tab. \ref{tab:ablation_vision_encorder}.

We found training ViT from scratch to be challenging (with only 22\% success rate), likely due to the limited amount data.
We also found training with frozen pretrained vision encoder to yield poor performance, which indicates that diffusion policy prefers different vision representation than what is offered in popular pretraining methods.
However, we found finetuning the pretrained vision encoder with a small learning rate (10x smaller vs diffusion policy network) gives the best performance overall. This is especially true for the CLIP-trained ViT-B/16, which reaches 98\% success rate with only 50 epochs of training.
Overall, the best performance across different architectures is not large, despite their significant theoretical capacity gap. We anticipate that their performance gap could be more pronounced on a complex task.

\begin{table}
\centering
\begin{tabular}{r|c|cc}
\toprule
Archicture \& & From & \multicolumn{2}{c}{Pretrained} \\
Prertain Datset& Scatch & frozen & finetuning \\
\midrule
Resnet18 (in21) & 0.94   & 0.58      & 0.92             \\
Resnet34 (in21)& 0.92   & 0.40      & 0.94             \\
ViT-base (clip)& 0.22   & 0.70      & 0.98             \\
\bottomrule
\end{tabular}
\caption{\textbf{Vision Encoder Comparison} All models are trained on the robomimic square (ph) task using CNN-based diffusion policy. Each model is trained for 500 epochs and evaluated every 50 epochs under 50 different environment initial conditions.}
\label{tab:ablation_vision_encorder} 
\vspace{-2mm}
\end{table}

\section{Realworld Evaluation}
We evaluated Diffusion Policy in the realworld performance on 4 tasks across 2 hardware setups -- with training data from different demonstrators for each setup. On the realworld Push-T task, we perform ablations examining Diffusion Policy on 2 architecture options and 3 visual encoder options; we also benchmarked against 2 baseline methods with both position-control and velocity-control action spaces. On all tasks, Diffusion Policy variants with both CNN backbones and end-to-end-trained visual encoders yielded the best performance. More details about the task setup and parameters may be found in supplemental materials.

\begin{table}[t]
\centering
\includegraphics[width=0.9\linewidth]{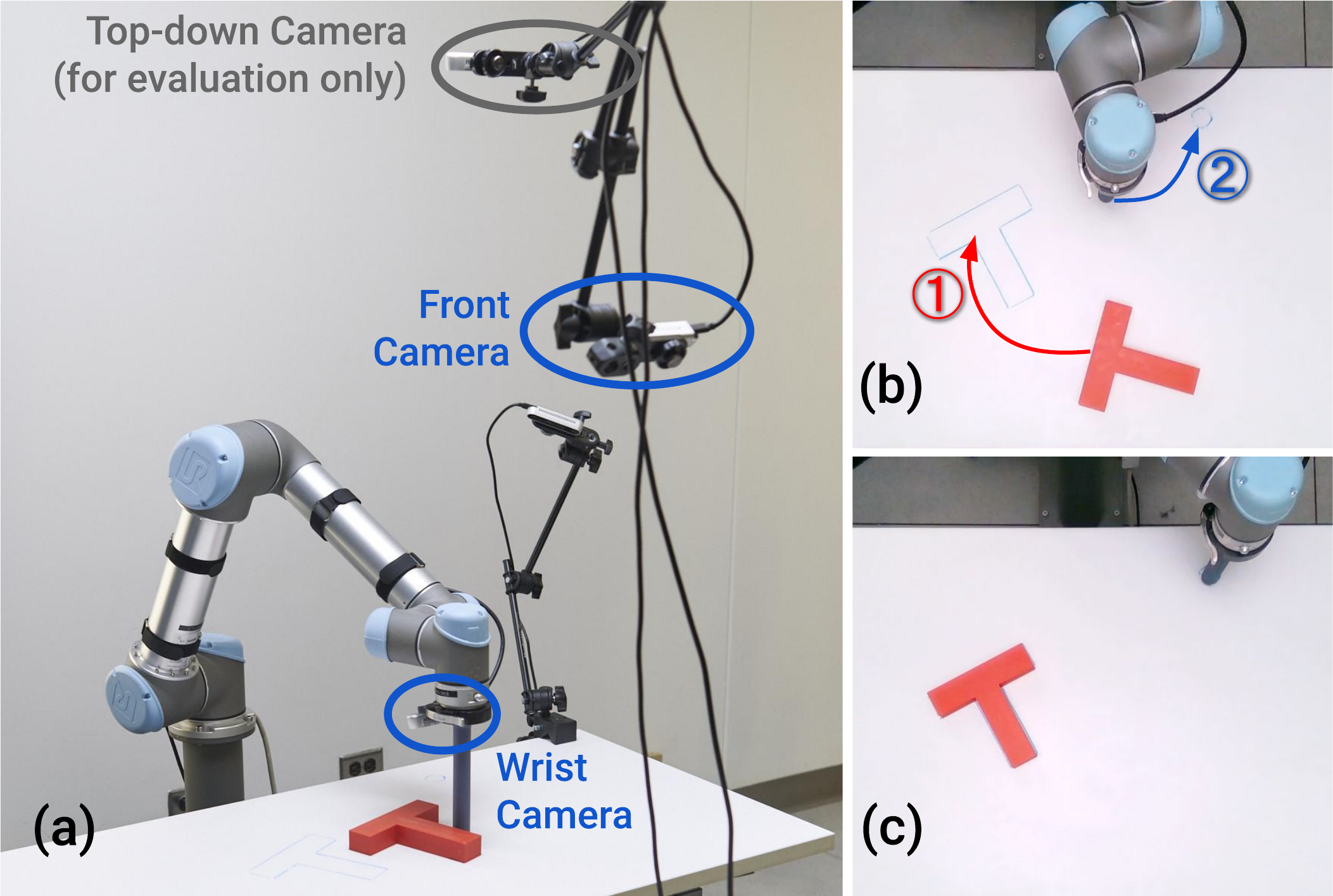}
% https://docs.google.com/drawings/d/1LZCbfrzt3Ww82h2LQ9rt8BGS1fJCw5TyY-AnGr1RfjE/edit
% \label{fig:real_pusht_setup}

\vspace{2mm}
\setlength\tabcolsep{1.2pt}
\small
\begin{tabular}{c|c|cc|cc|cccc}
\toprule
       & Human & \multicolumn{2}{c|}{IBC} & \multicolumn{2}{c|}{LSTM-GMM} & \multicolumn{4}{c}{Diffusion Policy}     \\
       & Demo  & pos        & vel        & pos          & vel         & T-E2E & ImgNet & R3M & E2E         \\
\midrule
IoU      & 0.84  & 0.14       & 0.19       & 0.24         & 0.25        & 0.53  & 0.24     & 0.66  & \textbf{0.80} \\
Succ\%   & 1.00  & 0.00       & 0.00       & 0.20         & 0.10        & 0.65  & 0.15     & 0.80  & \textbf{0.95} \\
Dur. & 20.3  & 56.3       & 41.6       & 47.3         & 51.7        & 57.5  & 55.8     & 31.7  & \textbf{22.9} \\
\bottomrule
% https://docs.google.com/spreadsheets/d/1TftzQuEmERMvSM4EpzPl7GZ2mw5Ym024wV2TPW9N6R4/edit#gid=577907863
\end{tabular}

\caption{\textbf{Realworld Push-T Experiment.} 
\label{tab:real_pusht}
a) Hardware setup.  
b) Illustration of the task. The robot needs to \textcircled{\raisebox{-0.9pt}{1}} precisely push the T-shaped block into the target region, \textbf{and} \textcircled{\raisebox{-0.9pt}{2}} move the end-effector to the end-zone. 
c) The ground truth end state used to calculate IoU metrics used in this table. Table: Success is defined by the end-state IoU greater than the minimum IoU in the demonstration dataset. Average episode duration presented in seconds. T-E2E stands for end-to-end trained Transformer-based Diffusion Policy}

\vspace{-4mm}
\end{table}

\begin{figure*}
\centering
\includegraphics[width=\linewidth]{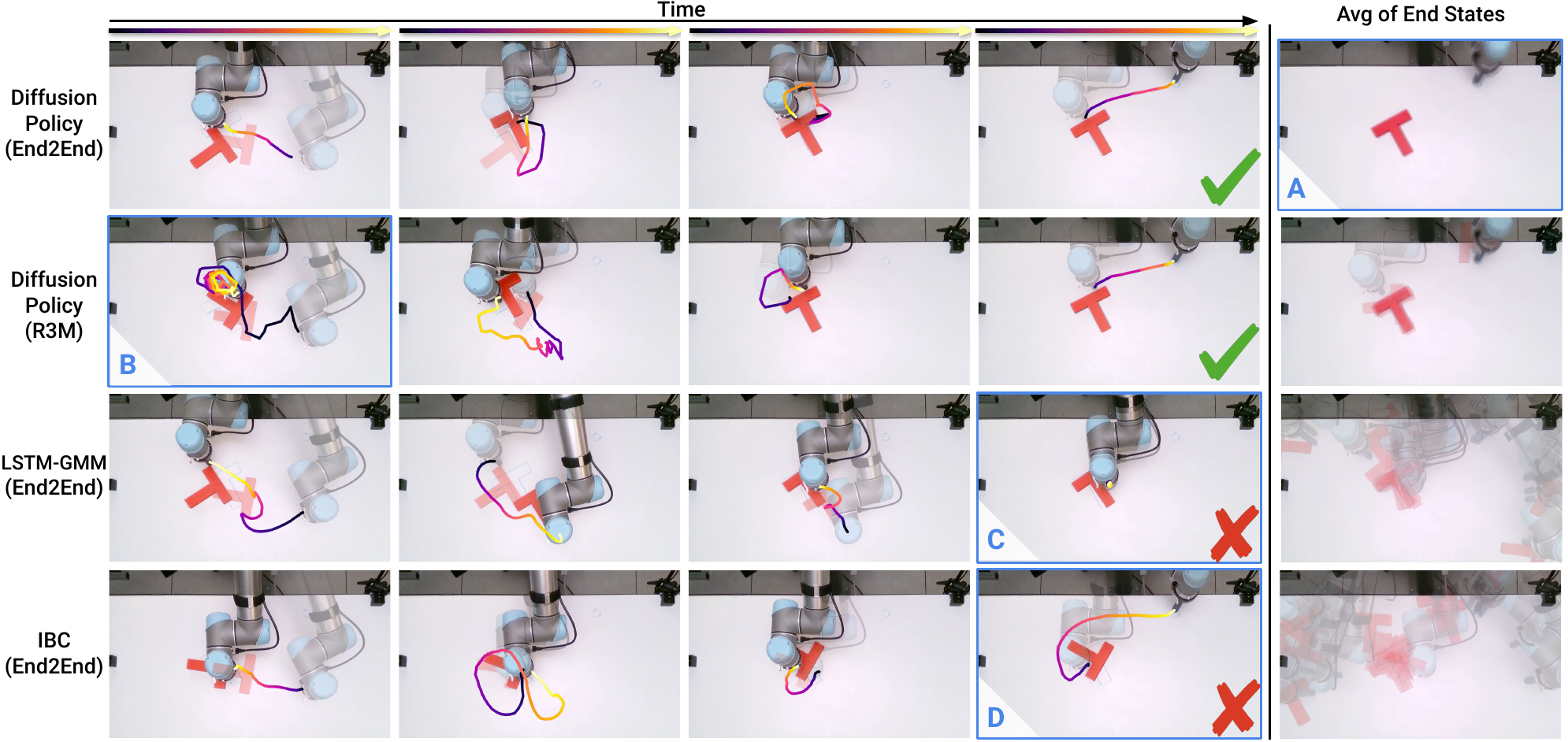}

% https://docs.google.com/drawings/d/1LY-oKJ32jSTlIpMnEnoyYGfd58Il9Zzf-YhFS8Axin0/edit
\caption{\textbf{Realworld Push-T Comparisons.} 
\label{fig:real_pusht_comparison}
Columns 1-4 show action trajectories based on key events. The last column shows averaged images of the end state. 
\textbf{A}: Diffusion policy (End2End) achieves more accurate and consistent end states.
\textbf{B}: Diffusion Policy (R3M) gets stuck initially but later recovers and finishes the task. 
\textbf{C}: LSTM-GMM fails to reach the end zone while adjusting the T block, blocking the eval camera view.
\textbf{D}: IBC prematurely ends the pushing stage.
}
\vspace{-2mm}
\end{figure*}

\subsection{Realworld Push-T Task}

Real-world Push-T is significantly harder than the simulated version due to 3 modifications: 1. The real-world Push-T task is \textbf{multi-stage}. It requires the robot to \textcircled{\raisebox{-0.9pt}{1}} push the T block into the target and then \textcircled{\raisebox{-0.9pt}{2}} move its end-effector into a designated end-zone to avoid occlusion. 2. The policy needs to make fine adjustments to make sure the T is fully in the goal region before heading to the end-zone, creating additional short-term multimodality. 3. The IoU metric is measured at the \textbf{last step} instead of taking the maximum over all steps. We threshold success rate by the minimum achieved IoU metric from the human demonstration dataset. Our UR5-based experiment setup is shown in Fig \ref{tab:real_pusht}. Diffusion Policy predicts robot commands at 10 Hz and these commands then linearly interpolated to 125 Hz for robot execution.

\textbf{Result Analysis.}
Diffusion Policy performed close to human level with 95\% success rate and 0.8 v.s. 0.84 average IoU, compared with the 0\% and 20\% success rate of best-performing IBC and LSTM-GMM variants. Fig \ref{fig:real_pusht_comparison} qualitatively illustrates the behavior for each method starting from the same initial condition. 
We observed that poor performance during the transition between stages is the most common failure case for the baseline method due to high multimodality during those sections and an ambiguous decision boundary. LSTM-GMM got stuck near the T block in 8 out of 20 evaluations (3rd row), while IBC prematurely left the T block in 6 out of 20 evaluations (4th row). 
We did not follow the common practice of removing \textbf{idle actions} from training data due to task requirements, which also contributed to LSTM and IBC's tendency to overfit on small actions and get stuck in this task. The results are best appreciated with videos in supplemental materials.

\textbf{End-to-end v.s. pre-trained vision encoders}
We tested Diffusion Policy with pre-trained vision encoders (ImageNet \cite{deng2009imagenet} and R3M\cite{nair2022r3m}), as seen in Tab. \ref{tab:real_pusht}. Diffusion Policy with R3M achieves an 80\% success rate but predicts jittery actions and is more likely to get stuck compared to the end-to-end trained version. Diffusion Policy with ImageNet showed less promising results with abrupt actions and poor performance. We found that end-to-end training is still the most effective way to incorporate visual observation into Diffusion Policy, and our best-performing models were all end-to-end trained.

\begin{figure}[t]
\centering
\includegraphics[width=\linewidth]{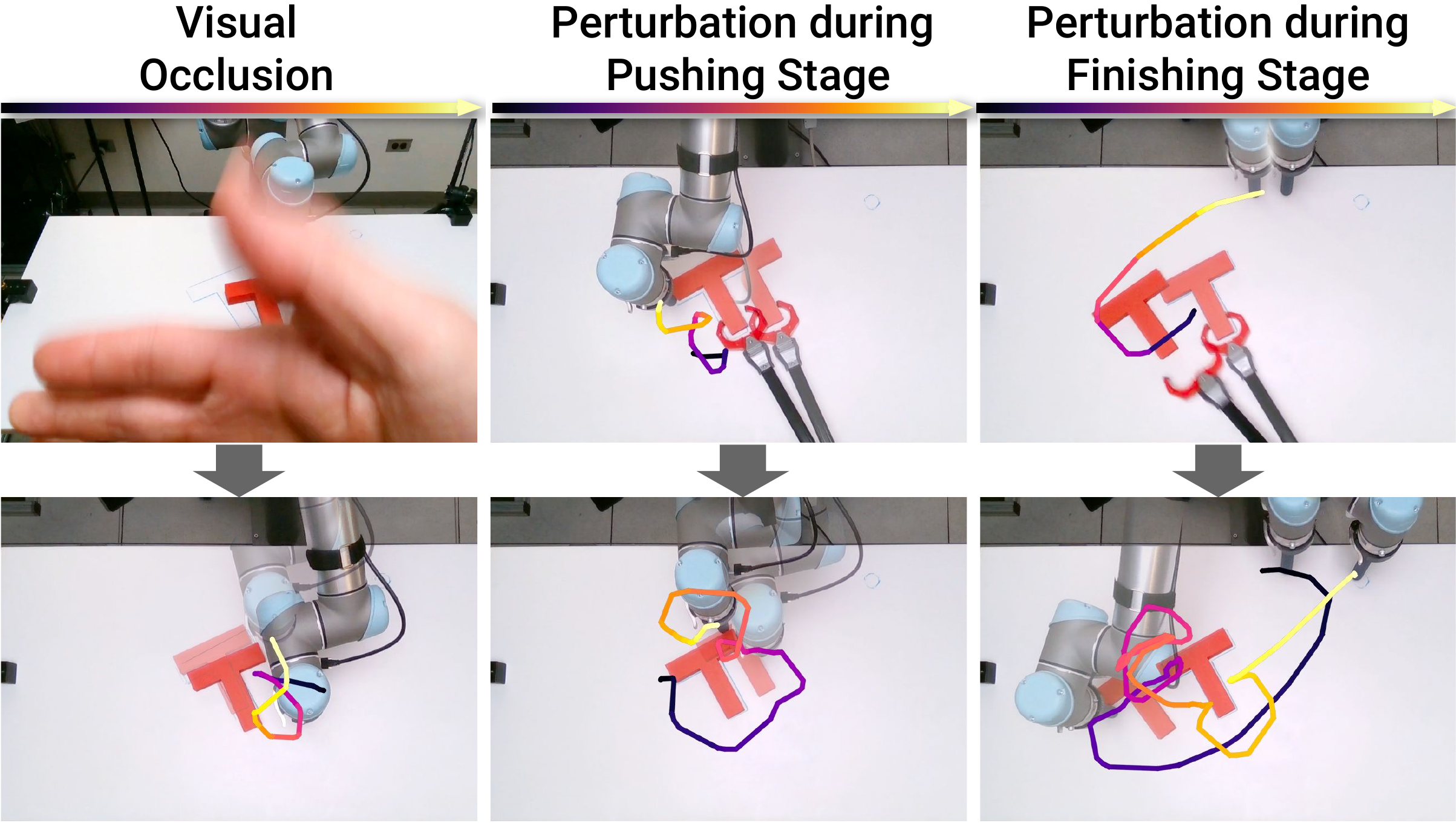}

% https://docs.google.com/drawings/d/13XwTyXJvdaa6HUrVMTcfdnoLmYsXkhqscgtLth0J-zk/edit
\caption{\textbf{Robustness Test for Diffusion Policy.} 
\label{fig:robustness}
\textbf{Left}: A waving hand in front of the camera for 3 seconds causes slight jitter, but the predicted actions still function as expected. 
% Left: Waving hand in front of the front camera for 3 seconds. The predicted actions are slightly jittery but still function as expected. 
\textbf{Middle}: Diffusion Policy immediately corrects shifted block position to the goal state during the pushing stage.
\textbf{Right}: Policy immediately aborts heading to the end zone, returning the block to goal state upon detecting block shift. This novel behavior was never demonstrated.
% Shifting the block while the policy has finished manipulating the block and is heading toward the end zone. The policy immediately aborts from moving to the end zone and moves the block back to the goal state. The behavior of aborting from heading to the end zone was never included in the demonstration. 
Please check the videos in the supplementary material. }

\vspace{-4mm}
\end{figure}

\textbf{Robustness against perturbation}
Diffusion Policy's robustness against visual and physical perturbations was evaluated in a separate episode from experiments in Tab \ref{tab:real_pusht}. As shown in Fig \ref{fig:robustness}, three types of perturbations are applied. 
1) The front camera was blocked for 3 secs by a waving hand (left column), but the diffusion policy, despite exhibiting some jitter, remained on-course and pushed the T block into position.
2) We shifted the T block while Diffusion Policy was making fine adjustments to the T block's position. Diffusion policy immediately re-planned to push from the opposite direction, negating the impact of perturbation. 
3) We moved the T block while the robot was en route to the end-zone after the first stage's completion. The Diffusion Policy immediately changed course to adjust the T block back to its target and then continued to the end-zone. This experiment indicates that Diffusion Policy may be able to \textbf{synthesize novel behavior} in response to unseen observations.
% In a separate evaluation episode from the experiment to generate Tab \ref{tab:real_pusht}, we examine Diffusion Policy's robustness against visual and physical perturbations. 
% As shown in Fig \ref{fig:robustness}, three types of perturbations are applied. 
% Shown in the left column, the front camera is blocked by a quickly waving hand for 3 seconds. Diffusion policy predicts action with increased jitteryness, but was able to remain on-course to push the T block into position, which demonstrates the robustness of end-to-end trained vision encoders despite being trained without any augmentations. 
% In the middle column, we shifted the T block while Diffusion Policy was fine-adjusting the T block's position. Diffusion policy immediately re-planed to push from the opposite direction, negating the impact of perturbation. 
% In the right column, while the robot has pushed the T block in to the target and is en route to the end-zone, we moved the T block out of the target. Diffusion Policy immediately changed course back to adjust the T block back to its target, before proceeding to the end-zone. There is no demonstration in the training dataset that exhibits the behavior of going back to T block pushing while previously en route to the end zone. 
% This experiment demonstrates that Diffusion Policy is able to \textbf{synthesize novel behavior} in the face of unseen observations.

\begin{figure}[t]
\centering
\includegraphics[width=0.9\linewidth]{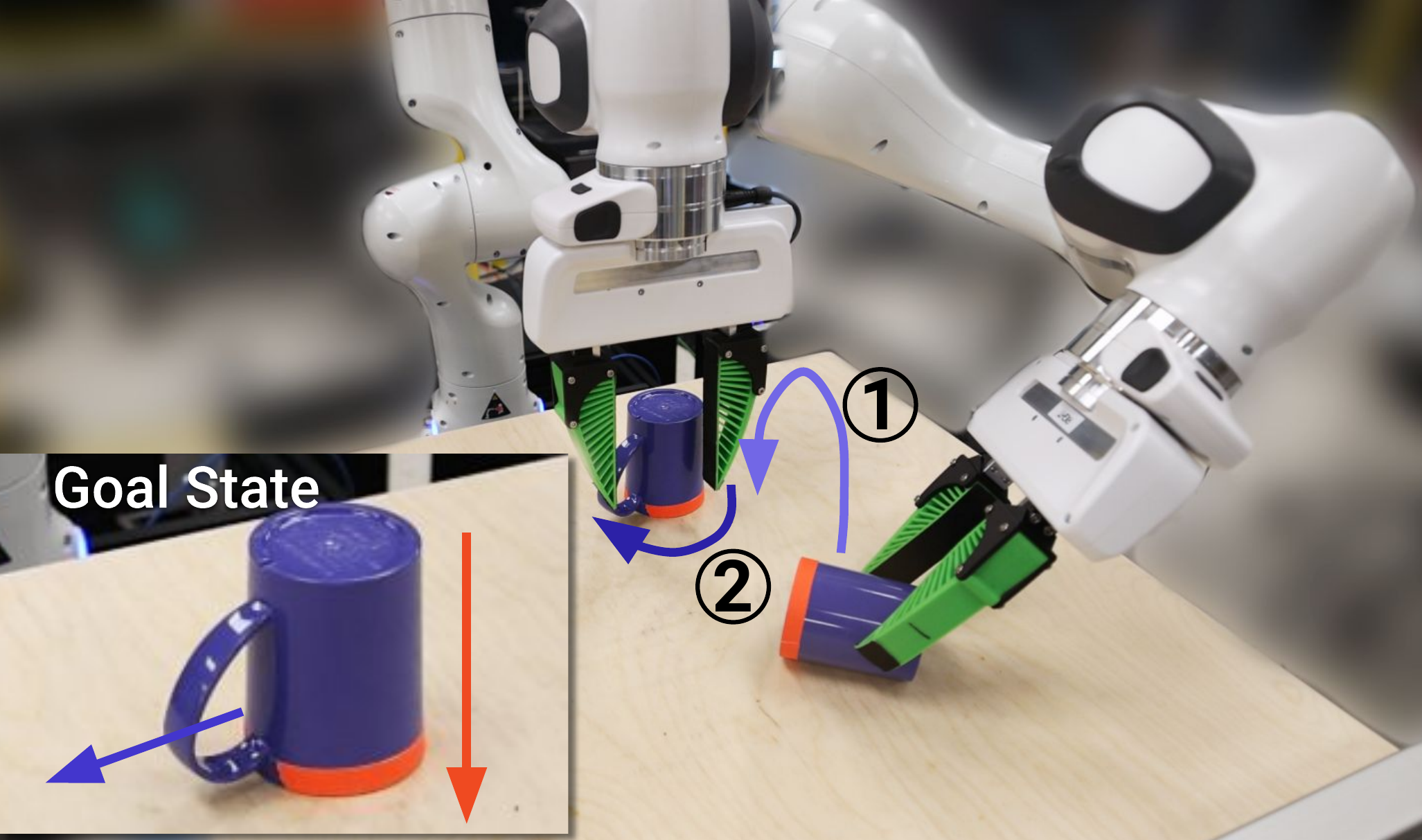}
% https://docs.google.com/drawings/d/169fC65dwdP1iOt87cCU2_9R8WrIZPQjq1YwE9EZejW4/edit

\vspace{1.5mm}
\begin{tabular}{c|c|c|c}
\toprule
        & Human & LSTM-GMM & Diffusion Policy \\
\midrule
Succ \% & 1.0   & 0.0      & 0.9             \\
\bottomrule
\end{tabular}

\caption{
\textbf{6DoF Mug Flipping Task.} 
\label{fig:mug_task}
The robot needs to 
\textcircled{\raisebox{-0.9pt}{1}} Pickup a randomly placed mug and place it lip down (marked orange).
\textcircled{\raisebox{-0.9pt}{2}} Rotate the mug such that its handle is pointing left.
}

\vspace{-4mm}
\end{figure}

\begin{figure}[t]
\centering
\includegraphics[width=\linewidth]{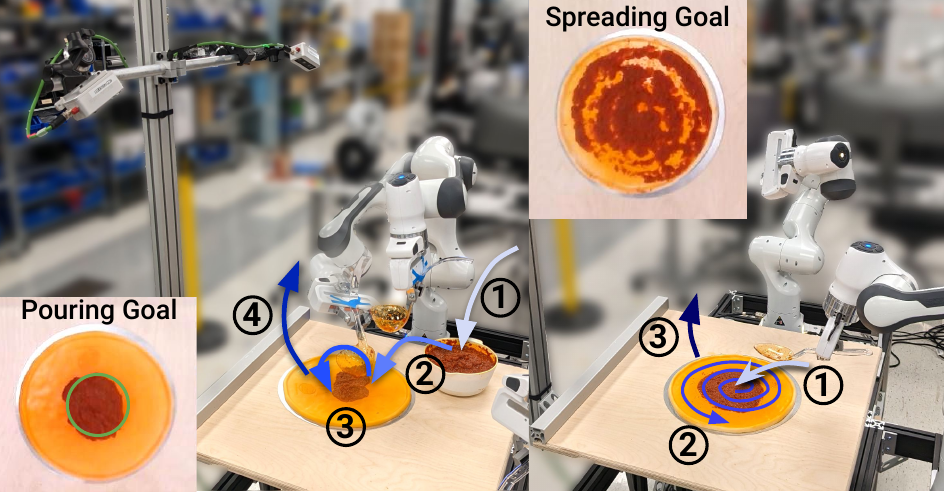}
% https://docs.google.com/drawings/d/1j7LKFAxZ2SXro3YbTC49KA1a7iuld-u5aY0oJHZzNUA/edit
% \includegraphics[width=0.5\linewidth]{figure/real_spreading.pdf}
% https://docs.google.com/drawings/d/1XDVS3JRA8forPWfl3t-7kVvTgDhtGlcqK0HsKyE-0Xc/edit

\vspace{1.5mm}

\small
\begin{tabular}{r|c|c|c|c}
\toprule
         & \multicolumn{2}{c|}{Pour} & \multicolumn{2}{c}{Spread}    \\
         & IoU & Succ & Coverage & Succ \% \\
\midrule
Human  & 0.79 &     1.00  & 0.79  &    1.00      \\
\midrule
LSTM-GMM & 0.06 & 0.00  &  0.27  & 0.00           \\
Diffusion Policy  &  \textbf{0.74} &\textbf{0.79}   & \textbf{0.77}  & \textbf{1.00}   \\
\bottomrule
\end{tabular}

\caption{\textbf{Realworld Sauce Manipulation. } 
% \todo{Update the table, double check the number} 
\label{fig:real_sauce_manipulation}
[Left] \textbf{6DoF pouring Task.} The robot needs to \textcircled{\raisebox{-0.9pt}{1}} dip the ladle to scoop sauce from the bowl, \textcircled{\raisebox{-0.9pt}{2}} approach the center of the pizza dough, \textcircled{\raisebox{-0.9pt}{3}} pour sauce, and \textcircled{\raisebox{-0.9pt}{4}} lift the ladle to finish the task.
[Right] \textbf{Periodic spreading Task} The robot needs to \textcircled{\raisebox{-0.9pt}{1}} approach the center of the sauce with a grasped spoon, \textcircled{\raisebox{-0.9pt}{2}} spread the sauce to cover pizza in a spiral pattern, and \textcircled{\raisebox{-0.9pt}{3}} lift the spoon to finish the task.
}

\vspace{-4mm}
\end{figure}

\subsection{Mug Flipping Task}
The mug flipping task is designed to test Diffusion Policy's ability to handle complex \textbf{3D rotations} while operating close to the hardware's kinematic limits.
The goal is to reorient a randomly placed mug to have \textcircled{\raisebox{-0.9pt}{1}} the lip facing down \textcircled{\raisebox{-0.9pt}{2}} the handle pointing left, as shown in Fig. \ref{fig:mug_task}.
% To generate the 250 demonstration episodes, the operator uses a mix strategies with multiple stages.
Depending on the mug's initial pose, the demonstrator might directly place the mug in desired orientation, or may use additional push of the handle to rotation the mug.
As a result, the demonstration dataset is highly multi-modal: grasp vs push, different types of grasps (forehand vs backhand) or local grasp adjustments (rotation around mug's principle axis), and are particularly challenging for baseline approaches to capture. 

\textbf{Result Analysis.} Diffusion policy is able to complete this task with 90\% success rate over 20 trials. The richness of captured behaviors is best appreciated with the video. Although never demonstrated, the policy is also able to sequence multiple pushes for handle alignment or regrasps for dropped mug when necessary. For comparison, we also train a LSTM-GMM policy trained with a subset of the same data. For 20 in-distribution initial conditions, the LSTM-GMM policy never aligns properly with respect to the mug, and fails to grasp in all trials.

\subsection{Sauce Pouring and Spreading}
The sauce pouring and spreading tasks are designed to test Diffusion Policy's ability to work with \textbf{non-rigid} objects, \textbf{6 Dof} action spaces, and \textbf{periodic} actions in real-world setups. Our Franka Panda setup and tasks are shown in Fig \ref{fig:real_sauce_manipulation}. The goal for the \textbf{6DoF pouring task} is to pour one full ladle of sauce onto the center of the pizza dough, with performance measured by IoU between the poured sauce mask and a nominal circle at the center of the pizza dough (illustrated by the green circle in Fig \ref{fig:real_sauce_manipulation}). 
The goal for the \textbf{periodic spreading task} is to spread sauce on pizza dough, with performance measured by sauce coverage. 
Variations across evaluation episodes come from random locations for the dough and the sauce bowl. 
The success rate is computed by thresholding with minimum human performance. 
Results are best viewed in supplemental videos. 
% For both tasks, we train with the same hyperparameters from the Pusth-T task and get successful policies on our first attempt.
Both tasks were trained with the same Push-T hyperparameters, and successful policies were achieved on the first attempt.

The sauce pouring task requires the robot to remain stationary for a period of time to fill the ladle with viscous tomato sauce. The resulting idle actions are known to be challenging for behavior cloning algorithms and therefore are often avoided or filtered out. Fine adjustments during pouring are necessary during sauce pouring to ensure coverage and to achieve the desired shape.

The demonstrated sauce-spreading strategy is inspired by the human chef technique, which requires both a long-horizon cyclic pattern to maximize coverage and short-horizon feedback for even distribution (since the tomato sauce used often drips out in lumps with unpredictable sizes). Periodic motions are known to be difficult to learn and therefore are often addressed by specialized action representations \cite{yang2022periodic}.
Both tasks require the policy to self-terminate by lifting the ladle/spoon.

% Due to the wide opening of the ladle and the lumpiness of the sauce, fine pose adjustments are necessary during sauce pouring to ensure the location and shape of the poured sauce. This task also has different phases with similar appearances, which can make policy learning challenging. Our sauce-spreading strategy is inspired by a common human chef technique. This strategy is cyclic in general, but it also requires fine adjustments of the spoon position relative to the edge of the sauce in order to push just the right amount of sauce outwards. This requires the policy to capture both the long-horizon cyclic pattern and short-horizon feedback. For both tasks, the policy needs to learn when to lift the tool to finish. 

% \textbf{Pouring Evaluation}
% Goal for this task is to pour one full ladle of sauce onto the center the pizza dough. Score is assigned by the IoU between the poured sauce mask and a nominal circle at the center of the pizza dough. Radius of the nominal circle is computed from the average poured sauce mask size from all the expert demonstrations. Since this task has more distinctive stages, we also assigned partial credits to roll outs that did not fully complete the task. 

% \textbf{Spreading Evaluation}
% Goal for this task is to evenly cover pizza dough with sauce. Amount of sauce is fixed during all training and evaluation runs. Score is assigned based on sauce coverage. 

\textbf{Result Analysis.}
Diffusion policy achieves close-to-human performance on both tasks, with coverage 0.74 vs 0.79 on pouring and 0.77 vs 0.79 on spreading. 
Diffusion policy reacted gracefully to external perturbations such as moving the pizza dough by hand during pouring and spreading. 
Results are best appreciated with videos in the supplemental material. 

LSTM-GMM performs poorly on both sauce pouring and spreading tasks. It failed to lift the ladle after successfully scooping sauce in 15 out of 20 of the pouring trials. When the ladle was successfully lifted, the sauce was poured off-centered. LSTM-GMM failed to self-terminate in all trials. We suspect LSTM-GMM's hidden state failed to capture sufficiently long history to distinguish between the ladle dipping and the lifting phases of the task. For sauce spreading, LSTM-GMM always lifts the spoon right after the start, and failed to make contact with the sauce in all 20 experiments. 
% We suspect the cyclic demonstrations make LSTM-GMM training hard.
\section{Realworld Bimanual Tasks}
\label{sec:eval_bimanual}
Beyond single arm setup, we further demonstrate Diffusion Policy on several challenging bimanual tasks. To enable bimanual tasks, the majority of effort was spent on extending our robot stack to support multi-arm teleopration and control. Diffusion Policy worked out of the box for these tasks without hyperparameter tuning. 

\subsection{Observation and Action Spaces}
The proprioceptive observation space is extended to include the poses of both end-effectors and the gripper widths of both grippers. We also extend the observation space to include the actual and desired values of these quantities.
The image observation space is comprised of two scene cameras and two wrist cameras, one attached to each arm.
The action space is extended to include the desired poses of both end-effectors and the desired gripper widths of both grippers.

\subsection{Teleoperation}
For these coordinated bimanual tasks, we found using 2 SpaceMouse simultaneously quite challenging for the demonstrator. Thus, we implemented two new teleoperation modes:
using a Meta Quest Pro VR device with two hand controllers, or
haptic-enabled control using 2
\href{https://www.haption.com/en/products-en/virtuose-6d-tao-en.html\#fa-download-downloads}{
Haption Virtuose\legalTM\ 6D HF TAO}
devices using bilateral position-position coupling as described succinctly in the haptics section of \citet{siciliano2008springer}. This coupling is performed between a Haption device and a Franka Panda arm.
More details on the controllers themselves may be found in Sec. \ref{sec:franka_setup}.
The following provides more details on each task and policy performance.  

\subsection{Bimanual Egg Beater}

\begin{figure}[t]
\centering
\includegraphics[width=\linewidth]{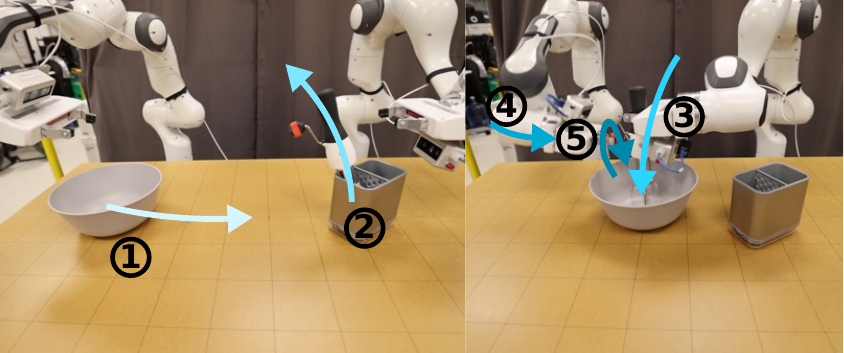}
\caption{\textbf{Bimanual Egg Beater Manipulation. } 
\label{fig:real_egg_beater}
The robot needs to
\textcircled{\raisebox{-0.9pt}{1}} push the bowl into position (only if too close to the left arm),
\textcircled{\raisebox{-0.9pt}{2}} approach and pick up the egg beater with the right arm,
\textcircled{\raisebox{-0.9pt}{3}} place the egg beater in the bowl,
\textcircled{\raisebox{-0.9pt}{4}} approach and grasp the egg beater crank handle, and
\textcircled{\raisebox{-0.9pt}{5}} turn the crank handle 3 or more times.
}
\vspace{-4mm}
\end{figure}

The bimanual egg beater task is illustrated and described in Fig. \ref{fig:real_egg_beater}, using a
\href{https://www.oxo.com/egg-beater.html}{OXO\legalTM Egg Beater} and a
\href{https://www.target.com/p/114oz-plastic-serving-bowl-jet-gray-room-essentials-8482/-/A-86701588}{Room Essentials\legalTM plastic bowl}. We chose this task to illustrate the importance of haptic feedback for teleoperating bimanual manipulation even for common daily life tasks such as coordinated tool use. Without haptic feedback, an expert was unable to successfully complete a single demonstration out of 10 trials. 5 failed due to robot pulling the crank handle off the egg beater; 3 failed due to robot losing grasp of the handle; 2 failed due to robot triggering torque limit. In contrast, the same operator could easily perform this task 10 out of 10 times with haptic feedback. Using haptic feedback made it possible for the demonstrations to be both quicker and higher quality than without feedback.

\textbf{Result Analysis.} Diffusion policy is able to complete this task with 55\% success rate over 20 trials, trained using 210 demonstrations. The primary failure modes for these were out-of-domain initial positioning of the egg beater, or missing the egg beater crank handle or losing grasp of it. The initial and final states for all rollouts are visualized in \ref{fig:egg_beater_ini} and \ref{fig:egg_beater_last}.

\subsection{Bimanual Mat Unrolling}

\begin{figure}[t]
\centering
\includegraphics[width=\linewidth]{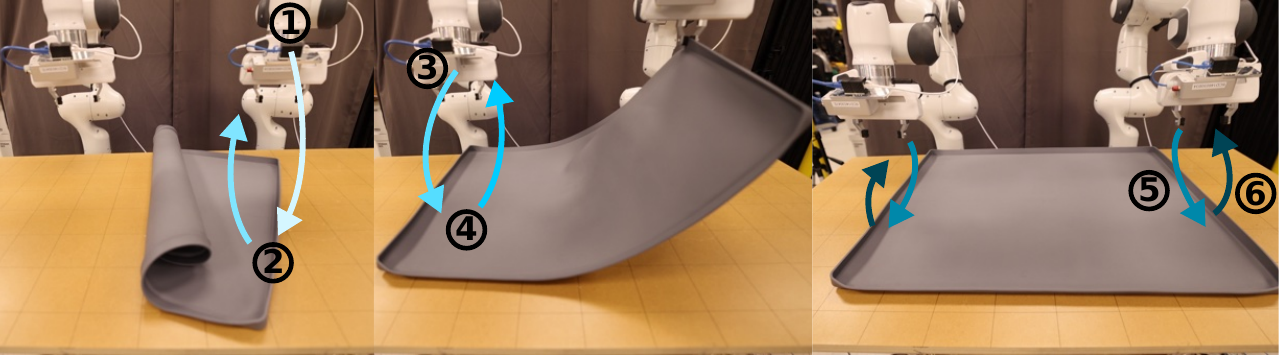}
\caption{\textbf{Bimanual Mat Unrolling. } 
\label{fig:real_unroll_mat}
The robot needs to
\textcircled{\raisebox{-0.9pt}{1}} pick up one side of the mat (if needed), using the left or right arm,
\textcircled{\raisebox{-0.9pt}{2}} lift and unroll the mat (if needed),
\textcircled{\raisebox{-0.9pt}{3}} ensure that both sides of the mat are grasped,
\textcircled{\raisebox{-0.9pt}{4}} lift the mat,
\textcircled{\raisebox{-0.9pt}{5}} place the mat oriented with the table, mostly centered, and
\textcircled{\raisebox{-0.9pt}{6}} release the mat.
}
\vspace{-4mm}
\end{figure}

The mat unrolling task is shown and described in Fig. \ref{fig:real_unroll_mat}, using a
\href{https://www.amazon.com/DogBuddy-Dog-Food-Mat-Waterproof/dp/B08GGDNB71}{XXL Dog Buddy\legalTM Dog Mat}.
This task was teleoperated using the VR setup, as it did not require rich haptic feedback to perform the task. We taught this skill to be omnidextrous, meaning it can unroll either to the left or right depending on the initial condition.

\textbf{Result Analysis.} Diffusion policy is able to complete this task with 75\% success rate over 20 trials, trained using 162 demonstrations. The primary failure modes for these were missed grasps during initial grasp of the mat, where the policy struggled to correct itself and thus got stuck repeating the same behavior. The initial and final states for all rollouts are visualized in \ref{fig:unroll_mat_ini} and \ref{fig:unroll_mat_last}.

\subsection{Bimanual Shirt Folding.}
\begin{figure}[t]
\centering
% TODO(eric): Couldn't get the PDF to fit under 50 MB
\includegraphics[width=\linewidth]{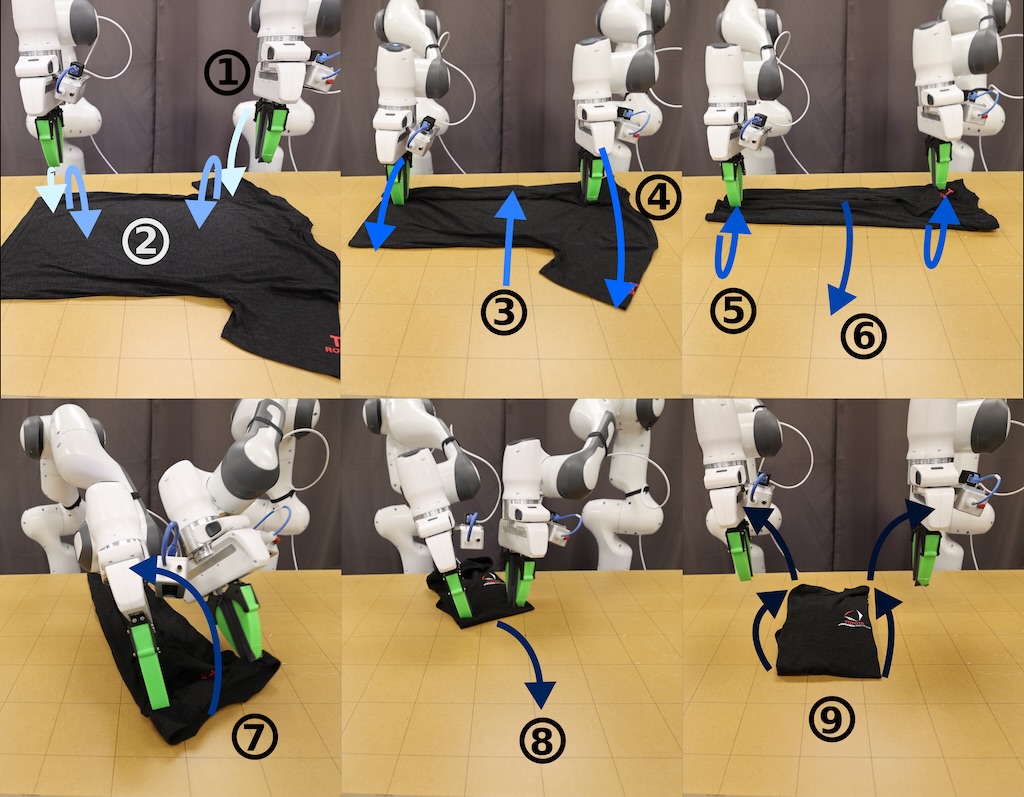}
\caption{\textbf{Bimanual Shirt Folding. } 
\label{fig:real_fold_shirt}
The robot needs to
\textcircled{\raisebox{-0.9pt}{1}} approach and grasp the closest sleeve with both arms,
\textcircled{\raisebox{-0.9pt}{2}} fold the sleeve and release,
\textcircled{\raisebox{-0.9pt}{3}} drag the shirt closer (if needed),
\textcircled{\raisebox{-0.9pt}{4}} approach and grasp the other sleeve with both arms,
\textcircled{\raisebox{-0.9pt}{5}} fold the sleeve and release,
\textcircled{\raisebox{-0.9pt}{6}} drag the shirt to a orientation for folding,
\textcircled{\raisebox{-0.9pt}{7}} grasp and fold the shirt in half by its collar,
\textcircled{\raisebox{-0.9pt}{8}} drag the shirt to the center, and
\textcircled{\raisebox{-0.9pt}{9}} smooth out the shirt and move the arms away.
}
\vspace{-4mm}
\end{figure}

The shirt folding task is described and illustrated in Fig. \ref{fig:real_fold_shirt}, using a short-sleeve T-shirt. This task was also teleoperated using the VR setup as it did not require rich
feedback to perform the task. Due to the kinematic and workspace constraints, this task is notably longer and can take up to nine discrete steps. The last few steps require both grippers to come very close towards each other. Having our mid-level controller explicitly handling collision avoidance was especially important for both teleoperation and policy rollout. 

\textbf{Result Analysis.} Diffusion policy is able to complete this task with 75\% success rate over 20 trials, trained using 284 demonstrations. The primary failure modes for these were missed grasps for initial folding (the sleeves and the color), and the policy being unable to stop adjusting the shirt at the end. The initial and final states for all rollouts are visualized in \ref{fig:fold_shirt_ini} and \ref{fig:fold_shirt_last}.

\section{Related Work}
Creating capable robots without requiring explicit programming of behaviors is a longstanding challenge in the field \cite{atkeson1997robot, argall2009survey, ravichandar2020recent}.
%In pursuit of this goal, imitation learning and behavior cloning  \cite{hussein2017imitation} have seen significant interest due to their appealing workflow. 
While conceptually simple, behavior cloning has shown surprising promise on an array of real-world robot tasks, including manipulation \cite{zhang2018deep, florence2019self, mandlekar2020learning, mandlekar2020iris, zeng2021transporter, rahmatizadeh2018vision, avigal2022speedfolding} and autonomous driving \cite{pomerleau1988alvinn, bojarski2016end}. Current behavior cloning approaches can be categorized into two groups, depending on the policy's structure.

%\section{Related Work}
% Creating capable robots without requiring explicit programming of behaviors is a longstanding challenge in the field \cite{atkeson1997robot, argall2009survey, ravichandar2020recent}.
% In pursuit of this goal, imitation learning \cite{hussein2017imitation} --- a paradigm where a robot learns to imitate demonstrated behavior --- has seen significant interest due to its appealing workflow of directly teaching a robot by example. A commonly employed form of imitation learning is behavior cloning, a supervised and model-free approach where the learner is provided paired $\langle observation, action\rangle$ data during training and learns to predict actions conditioned on observations at inference time. While conceptually straightforward, behavior cloning has shown surprising promise on an array of real-world robot tasks, including \todo{task list and citations here for behavior cloning works we want to cite}. Current behavior cloning approaches can be categorized into two groups, depending on the structure of the learned policy being employed.

\textbf{Explicit Policy.}
The simplest form of explicit policies maps from world state or observation directly to action \cite{pomerleau1988alvinn, zhang2018deep, florence2019self, ross2011reduction, toyer2020magical, rahmatizadeh2018vision, bojarski2016end}. They can be supervised with a direct regression loss and have efficient inference time with one forward pass. Unfortunately, this type of policy is not suitable for modeling multi-modal demonstrated behavior, 
% since the L2 regression loss implies a Gaussian distribution for the predicted value \cite{mathieu2016mse}, 
and struggles with high-precision tasks \cite{ibc}.
% The simplest form of explicit policies maps from world state or observation directly to action \todo{citations, probably including bcrnn here}. 
% This formulation has several appealing attributes. 
% First, it is extremely straightforward to supervise such an approach during behavior cloning with a standard direct regression loss on predicted action error with respect to demonstrated behavior. 
% Explicit policies also benefit from inference-time efficiency. Because actions are directly predicted by the policy, only a single forward pass is needed at each timestep to produce an action. 
% Unfortunately, this type of policy is not suitable for modeling multi-modal demonstrated behavior and struggles with high-precision tasks \cite{ibc}.
A popular approach to model multimodal action distributions while maintaining the simplicity of direction action mapping is convert the regression task into classification by discretizing the action space \cite{zeng2021transporter, wu2020spatial, avigal2022speedfolding}. 
% However, the behavior temporal complexity of this approach is often limited, since the number of bins needed grows exponentially with increasing dimensionality.
However, the number of bins needed to approximate a continuous action space grows exponentially with increasing dimensionality. 
% Therefore, methods with discretized action space usually rely on hand-crafted action primitives \cite{zeng2021transporter, avigal2022speedfolding} in order to scale to complex tasks.
Another approach is to combine Categorical and Gaussian distributions to represent continuous multimodal distributions via the use of MDNs \cite{bishop1994mixture, robomimic} or clustering with offset prediction \cite{bet, sharma2018multiple}. Nevertheless, these models tend to be sensitive to hyperparameter tuning, exhibit mode collapse, and are still limited in their ability to express high-precision behavior \cite{ibc}.

% An extension to directly predicting actions is to instead predict explicit action distributions from which actions may be sampled \todo{Anything predicting unimodal distributions we should cite?}. If these predicted distributions admit multimodality, such as mixture models \todo{cite mixture density networks paper}, it is possible for explicit policies to model multimodal behavior.
% \todo{Talk about the limitation of discritized action space}
% Nevertheless, these models tend to be sensitive to hyperparmeter tuning, exhibit mode collapse during training, and are still limited in their ability to express high-precision behavior \cite{ibc}.

%\ben{Anything else we want to say about BCRNN?}

\textbf{Implicit Policy.}
Implicit policies \citep{ibc, jarrett2020strictly} define distributions over actions by using Energy-Based Models (EBMs) \citep{lecun06atutorial, du2019implicit, dai2019exponential, grathwohl2020stein, du2020improved}. 
In this setting, each action is assigned an energy value, with action prediction corresponding to the optimization problem of finding a minimal energy action. Since different actions may be assigned low energies, implicit policies naturally represent multi-modal distributions. However, existing implicit policies \citep{ibc} are unstable to train due to the necessity of drawing negative samples when computing the underlying Info-NCE loss. %Diffusion policy may be seen as a stable approach to train implicit policies, where we remove the necessity of drawing negative samples by instead directly learning gradients of the action energy landscape using denoising.  

% \crossmark

%\yilun{Would be good add additional works that build on top of IBC or use IBC for robotic control -- I'm not that familiar with those}

% \begin{table}[h]
%     \centering
%     \setlength\tabcolsep{5 pt}
%     \begin{tabular}{l|ccccc}
%     \toprule
%      %& IBC\cite{ibc} & LSTM-GMM\cite{robomimic} & BET\cite{bet} & Diffuser\cite{janner2022diffuser} & Ours\\
%      & \cite{ibc} & \cite{robomimic} & \cite{bet} & \cite{janner2022diffuser} & Ours\\
%     \midrule
%     Multimodal          & \cmark    & \cmark    & \cmark    & \cmark    & \cmark\\
%     Any Distribution    & \cmark    & \xmark    & \xmark    & \cmark    & \cmark\\
%     Stable training     & \xmark    & \cmark    & \cmark    & \cmark    & \cmark\\
%     High-dim action     & \xmark    & \xmark    & \cmark    & \cmark    & \cmark\\
%     Position control    & \cmark    & \xmark    & \xmark    & \cmark    & \cmark\\
%     Closed-loop control & \cmark    & \cmark    & \cmark    & \xmark    & \cmark\\
%     Visual input        & \cmark    & \cmark    & \cmark    & \xmark    & \cmark\\
%     \bottomrule
%     \end{tabular}
%     \caption{Add a table to contrast the capability? }
%     \label{tab:my_label}
% \end{table}

\textbf{Diffusion Models.}
Diffusion models are probabilistic generative models that iteratively refine randomly sampled noise into draws from an underlying distribution. They can also be conceptually understood as learning the gradient field of an implicit action score and then optimizing that gradient during inference.
Diffusion models \citep{sohldickstein2015nonequilibrium, ho2020denoising} have recently been applied to solve various different control tasks \citep{janner2022diffuser, urain2022se, ajay2022conditional}.
% more diffusion work can be added here 

In particular, \citet{janner2022diffuser} and \citet{huang2023diffusion} explore how diffusion models may be  used in the context of planning and infer a trajectory of actions that may be executed in a given environment. 
In the context of Reinforcement Learning, \citet{wang2022diffusion} use diffusion model for policy representation and regularization with state-based observations.
In contrast, in this work, we explore how diffusion models may instead be effectively applied in the context of behavioral cloning for effective visuomotor control policy. 
To construct effective visuomotor control policies, we propose to combine DDPM's ability to predict high-dimensional action squences with closed-loop control, as well as a new transformer architecture for action diffusion and a manner to integrate visual inputs into the action diffusion model.

\citet{wang2023diffusion} explore how diffusion models learned from expert demonstrations can be used to augment classical explicit polices without directly taking advantage of diffusion models as policy representation.

Concurrent to us, \citet{pearce2023imitating}, \citet{reuss2023goal} and \citet{hansen2023idql} has conducted a complimentary analysis of diffusion-based policies in simulated environments. While they focus more on effective sampling strategies, leveraging classifier-free guidance for goal-conditioning as well as applications in Reinforcement Learning, and we focus on effective action spaces, our empirical findings largely concur in the simulated regime.  In addition, our extensive real-world experiments provide strong evidence for the importance of a receding-horizon prediction scheme, the careful choice between velocity and position control, and the necessity of optimization for real-time inference and other critical design decisions for a physical robot system.

% \section{What Doesn't Work}
% Diffusion + BatchNorm (EMA, n\_obs\_steps).
% Shared vision encoder for all views (end-to-end).

\section{Limitations and Future Work}
% Although we have demonstrated diffusion policy’s effectiveness with extensive evaluations in both simulation and real-world systems, there are still several limitations that can be improved in future works. 
% First, our current implementation of Diffusion Policy inherits limitations from Behavior Cloning, such as the rescued performance when trained with sub-optimal demonstration data. Future works could apply Diffusion Policy to other paradigms such as Offline Reinforcement Learning to take advantage of suboptimal and negative data. 
% Second, Diffusion Policy incurs relatively high computational cost and inference latency compared to simpler methods such as LSTM-GMM. This is partially mitigated by our action sequence prediction approach however, may not be sufficient for tasks requiring high rate control. Future works could take advantage of latest advancement in diffusion model acceleration methods such as new noise schedules, inference solvers as well as consistency models to reduce number of inference steps required.

Although we have demonstrated the effectiveness of diffusion policy in both simulation and real-world systems, there are limitations that future work can improve. 
First, our implementation inherits limitations from behavior cloning, such as suboptimal performance with inadequate demonstration data. Diffusion policy can be applied to other paradigms, such as reinforcement learning \cite{wang2023diffusion,hansen2023idql}, to take advantage of suboptimal and negative data. 
Second, diffusion policy has higher computational costs and inference latency compared to simpler methods like LSTM-GMM. Our action sequence prediction approach partially mitigates this issue, but may not suffice for tasks requiring high rate control. Future work can exploit the latest advancements in diffusion model acceleration methods to reduce the number of inference steps required, such as new noise schedules \cite{chen2023importance}, inference solvers \cite{karras2022elucidating}, and consistency models \cite{song2023consistency}.

\section{Conclusion}
%Systematic studies on the characteristic of diffusion policy and demonstrate its effectiveness in the context of robot manipulation and behavior clone. 
%Through these practical recommendations on the key design decisions for applying diffusion policy to  different robotics tasks.

%\ben{Taking a stab at fleshing this out, feel free to modify:} 
In this work, we assess the feasibility of diffusion-based policies for robot behaviors. Through a comprehensive evaluation of 15 tasks in simulation and the real world, we demonstrate that diffusion-based visuomotor policies consistently and definitively outperform existing methods while also being stable and easy to train. Our results also highlight critical design factors, including receding-horizon action prediction, end-effector position control, and efficient visual conditioning, that is crucial for unlocking the full potential of diffusion-based policies. While many factors affect the ultimate quality of behavior-cloned policies --- including the quality and quantity of demonstrations, the physical capabilities of the robot, the policy architecture, and the pretraining regime used --- our experimental results strongly indicate that policy structure poses a significant performance bottleneck during behavior cloning. We hope that this work drives further exploration in the field into diffusion-based policies and highlights the importance of considering all aspects of the behavior cloning process beyond just the data used for policy training.

\section{Acknowledgement}
% This work was supported in part by  NSF Awards 2037101, 2132519, 2037101, and Toyota Research Institute. We would like to thank Google for the UR5 robot hardware. The views and conclusions contained herein are those of the authors and should not be interpreted as necessarily representing the official policies, either expressed or implied, of the sponsors.

We'd like to thank Naveen Kuppuswamy, Hongkai Dai, Aykut Önol, Terry Suh, Tao Pang, Huy Ha, Samir Gadre, Kevin Zakka and Brandon Amos for their thoughtful discussions. We thank Jarod Wilson for 3D printing support and Huy Ha for photography and lighting advice. We thank Xiang Li for discovering the bug in our evaluation code on GitHub.

% \todo{list the people we want to thank?}

% \begin{acks} 
% We would like to thank Zhenjia Xu, Huy Ha, Dale McConachie, Naveen Kuppuswamy for their helpful feedback and fruitful discussions.
% \end{acks}

\begin{funding}
This work was supported by the Toyota Research Institute, NSF CMMI-2037101 and NSF IIS-2132519. We would like to thank Google for the UR5 robot hardware. The views and conclusions contained herein are those of the authors and should not be interpreted as necessarily representing the official policies, either expressed or implied, of the sponsors.
\end{funding}

%%Harvard (name/date)
\bibliographystyle{SageH}
%%Vancouver (numbered)
%\bibliographystyle{SageV}
% \bibliographystyle{plainnat}
\bibliography{references}

\appendix
\section{Diffusion Policy Implementation Details}

\subsection{Normalization}
Properly normalizing action data is critical to achieve best performance for Diffusion Policy. 
Scaling the min and max of each action dimension independently to $[-1,1]$ works well for most tasks. 
Since DDPMs clip prediction to $[-1,1]$ at each iteration to ensure stability, the common zero-mean unit-variance normalization will cause some region of the action space to be inaccessible. 
When the data variance is small (e.g., near constant value), shift the data to zero-mean without scaling to prevent numerical issues.
We leave action dimensions corresponding to rotation representations (e.g. Quaternion) unchanged.

\subsection{Rotation Representation}
For all environments with velocity control action space, we followed the standard practice \cite{robomimic} to use 3D axis-angle representation for the rotation component of action. Since velocity action commands are usually close to 0, the singularity and discontinuity of the axis-angle representation don't usually cause problems. We used the 6D rotation representation proposed in \citet{zhou2019continuity} for all environments (real-world and simulation) with positional control action space. 

\subsection{Image Augmentation}
Following \citet{robomimic}, we employed random crop augmentation during training. The crop size for each task is indicated in Tab. \ref{tab:hparam_cnn}. During inference, we take a static center crop with the same size.

\subsection{Hyperparameters}
Hyerparameters used for Diffusion Policy on both simulation and realworld benchmarks are shown in Tab. \ref{tab:hparam_cnn} and Tab. \ref{tab:hparam_transformer}. Since the Block Push task uses a Markovian scripted oracle policy to generate demonstration data, we found its optimal hyper parameter for observation and action horizon to be very different from other tasks with human teleop demostrations.

We found that the optimal hyperparameters for CNN-based Diffusion Policy are consistent across tasks. In contrast, transformer-based Diffusion Policy's optimal attention dropout rate and weight decay varies greatly across different tasks. During tuning, we found increasing the number of parameters in CNN-based Diffusion Policy always improves performance, therefore the optimal model size is limited by the available compute and memory capacity. On the other hand, increasing model size for transformer-based Diffusion Policy (in particular number of layers) hurts performance sometimes. For CNN-based Diffusion Policy, We found using FiLM conditioning to pass-in observations is better than impainting on all tasks except Push-T. Performance reported for DiffusionPolicy-C on Push-T in Tab. \ref{tab:table_low_dim} used impaiting instead of FiLM.

On simulation benchmarks, we used the iDDPM algorithm \cite{nichol2021improved} with the same 100 denoising diffusion iterations for both training and inference. We used DDIM \cite{song2021ddim} on realworld benchmarks to reduce the inference denoising iterations to 16 therefore reducing inference latency.

We used batch size of 256 for all state-based experiments and 64 for all image-based experiments. For learning-rate scheduling, we used cosine schedule with linear warmup. CNN-based Diffusion Policy is warmed up for 500 steps while Transformer-based Diffusion Policy is warmed up for 1000 steps.

\begin{table*}
\centering
\setlength\tabcolsep{3 pt}
\small
\begin{tabular}{l|llllllllllll}
\toprule
           % & Ctrl & To & Ta & Tp & ImgRes    & CropRes   & \#D-params & \#V-params & Lr   & WDecay & D-Iters Train & D-Iters Eval \\
\textbf{H-Param} & \textbf{Ctrl} & \textbf{To} & \textbf{Ta} & \textbf{Tp} & \textbf{ImgRes} & \textbf{CropRes} & \textbf{\#D-Params} & \textbf{\#V-Params} & \textbf{Lr} & \textbf{WDecay} & \textbf{D-Iters Train} & \textbf{D-Iters Eval} \\

\midrule
Lift       & Pos  & 2  & 8  & 16 & 2x84x84   & 2x76x76   & 256      & 22       & 1e-4 & 1e-6   & 100           & 100          \\
Can        & Pos  & 2  & 8  & 16 & 2x84x84   & 2x76x76   & 256      & 22       & 1e-4 & 1e-6   & 100           & 100          \\
Square     & Pos  & 2  & 8  & 16 & 2x84x84   & 2x76x76   & 256      & 22       & 1e-4 & 1e-6   & 100           & 100          \\
Transport  & Pos  & 2  & 8  & 16 & 4x84x85   & 4x76x76   & 264      & 45       & 1e-4 & 1e-6   & 100           & 100          \\
ToolHang   & Pos  & 2  & 8  & 16 & 2x240x240 & 2x216x216 & 256      & 22       & 1e-4 & 1e-6   & 100           & 100          \\
Push-T     & Pos  & 2  & 8  & 16 & 1x96x96   & 1x84x84   & 256      & 22       & 1e-4 & 1e-6   & 100           & 100          \\
Block Push & Pos  & 3  & 1  & 12 & N/A       & N/A       & 256      & 0        & 1e-4 & 1e-6   & 100           & 100          \\
Kitchen    & Pos  & 2  & 8  & 16 & N/A       & N/A       & 256      & 0        & 1e-4 & 1e-6   & 100           & 100          \\
\midrule
Real Push-T     & Pos  & 2  & 6  & 16 & 2x320x240 & 2x288x216 & 67       & 22       & 1e-4 & 1e-6   & 100           & 16           \\
Real Pour       & Pos  & 2  & 8  & 16 & 2x320x240 & 2x288x216 & 67       & 22       & 1e-4 & 1e-6   & 100           & 16           \\
Real Spread     & Pos  & 2  & 8  & 16 & 2x320x240 & 2x288x216 & 67       & 22       & 1e-4 & 1e-6   & 100           & 16           \\
Real Mug Flip   & Pos  & 2  & 8  & 16 & 2x320x240 & 2x288x216 & 67       & 22       & 1e-4 & 1e-6   & 100           & 16           \\
\bottomrule
\end{tabular}
\caption{
\textbf{Hyperparameters for CNN-based Diffusion Policy}
\label{tab:hparam_cnn}
Ctrl: position or velocity control 
To: observation horizon 
Ta: action horizon 
Tp: action prediction horizon 
ImgRes: environment observation resolution (Camera views x W x H) 
CropRes: random crop resolution 
\#D-Params: diffusion network number of parameters in millions 
\#V-Params: vision encoder number of parameters in millions 
Lr: learining rate 
WDecay: weight decay
D-Iters Train: number of training diffusion iterations
D-Iters Eval: number of inference diffusion iterations (enabled by DDIM \cite{song2021ddim})
}

\vspace{4mm}
\centering
\setlength\tabcolsep{2.1 pt}
\begin{tabular}{l|lllllllllllll}
\toprule
           % & Ctrl & To & Ta & Tp & \#D-params & \#V-params & \#Layers & Emb Dim & Attn Dropout & Lr   & WDecay & D-Iters Train & D-Iters Eval \\
\textbf{H-Param} & \textbf{Ctrl} & \textbf{To} & \textbf{Ta} & \textbf{Tp} & \textbf{\#D-params} & \textbf{\#V-params} & \textbf{\#Layers} & \textbf{Emb Dim} & \textbf{Attn Drp} & \textbf{Lr} & \textbf{WDecay} & \textbf{D-Iters Train} & \textbf{D-Iters Eval} \\

\midrule
Lift       & Pos  & 2  & 8  & 10 & 9        & 22       & 8      & 256     & 0.3          & 1e-4 & 1e-3   & 100           & 100          \\
Can        & Pos  & 2  & 8  & 10 & 9        & 22       & 8      & 256     & 0.3          & 1e-4 & 1e-3   & 100           & 100          \\
Square     & Pos  & 2  & 8  & 10 & 9        & 22       & 8      & 256     & 0.3          & 1e-4 & 1e-3   & 100           & 100          \\
Transport  & Pos  & 2  & 8  & 10 & 9        & 45       & 8      & 256     & 0.3          & 1e-4 & 1e-3   & 100           & 100          \\
ToolHang   & Pos  & 2  & 8  & 10 & 9        & 22       & 8      & 256     & 0.3          & 1e-4 & 1e-3   & 100           & 100          \\
Push-T     & Pos  & 2  & 8  & 16 & 9        & 22       & 8      & 256     & 0.01         & 1e-4 & 1e-1   & 100           & 100          \\
Block Push & Vel  & 3  & 1  & 5  & 9        & 0        & 8      & 256     & 0.3          & 1e-4 & 1e-3   & 100           & 100          \\
Kitchen    & Pos  & 4  & 8  & 16 & 80       & 0        & 8      & 768     & 0.1          & 1e-4 & 1e-3   & 100           & 100          \\
\midrule
Real Push-T     & Pos  & 2  & 6  & 16 & 80      & 22       & 8      & 768     & 0.3          & 1e-4 & 1e-3   & 100           & 16           \\
\bottomrule
\end{tabular}
\caption{
\textbf{Hyperparameters for Transformer-based Diffusion Policy}
\label{tab:hparam_transformer}
Ctrl: position or velocity control 
To: observation horizon 
Ta: action horizon 
Tp: action prediction horizon 
\#D-Params: diffusion network number of parameters in millions 
\#V-Params: vision encoder number of parameters in millions 
Emb Dim: transformer token embedding dimension
Attn Drp: transformer attention dropout probability
Lr: learining rate 
WDecay: weight decay (for transformer only)
D-Iters Train: number of training diffusion iterations
D-Iters Eval: number of inference diffusion iterations (enabled by DDIM \cite{song2021ddim})
}
\vspace{-5mm}
\end{table*}

\begin{figure}
\centering
\includegraphics[width=\linewidth]{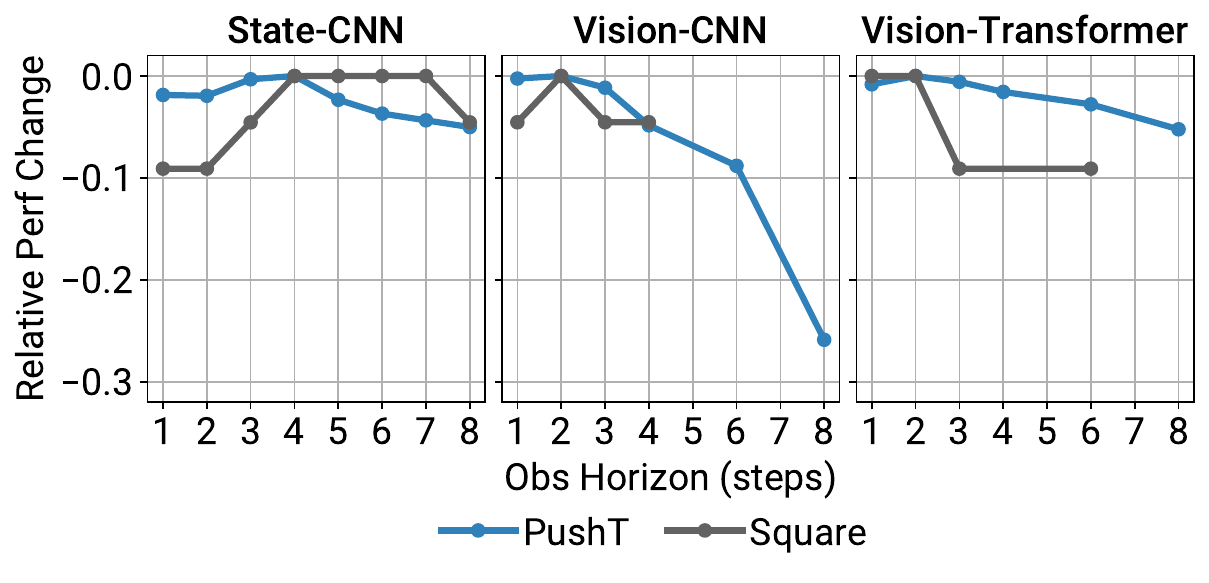}

\vspace{-2mm}
\caption{
\textbf{Observation Horizon Ablation Study.}
\label{fig:obs_horizon_ablation}
State-based Diffusion Policy is not sensitive to observation horizon.
Vision-based Diffusion Policy prefers low but $>1$ observation horizon, with $2$ being a good compromise for most tasks.
}
\vspace{-3mm}
\end{figure}

\begin{figure}
\centering
\includegraphics[width=\linewidth]{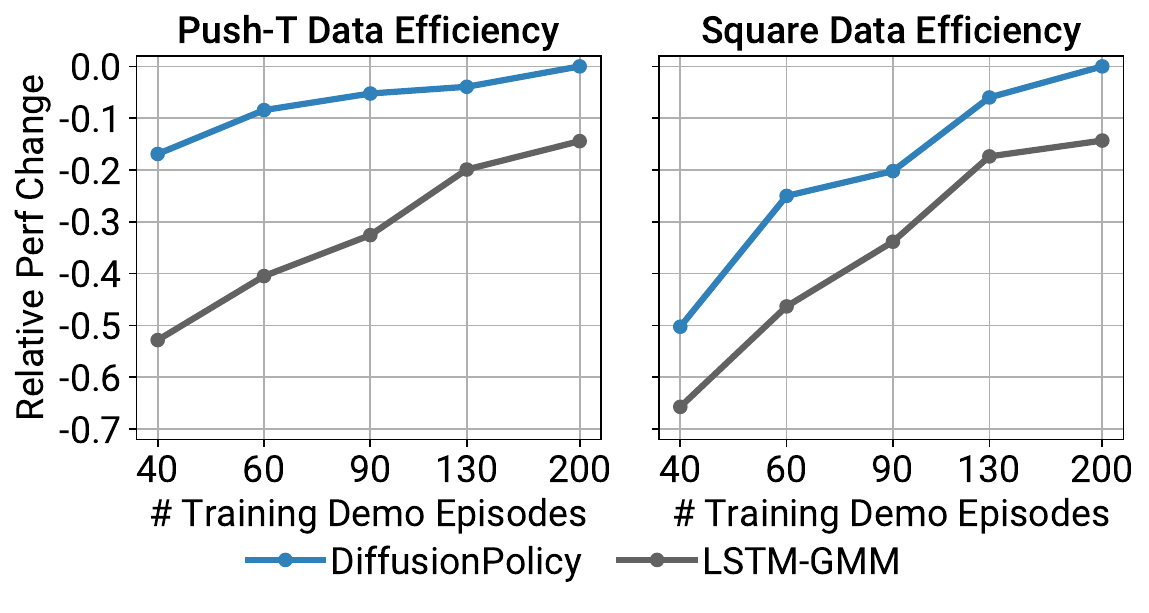}

\vspace{-2mm}
\caption{
\textbf{Data Efficiency Ablation Study.}
\label{fig:data_efficiency}
Diffusion Policy outperforms LSTM-GMM \cite{robomimic} at every training dataset size.
}
\vspace{-5mm}
\end{figure}

\subsection{Data Efficiency}
We found Diffusion Policy to outperform LSTM-GMM \cite{robomimic} at every training dataset size, as shown in Fig. \ref{fig:data_efficiency}.

\section{Additional Ablation Results}
\subsection{Observation Horizon}
We found state-based Diffusion Policy to be insensitive to observation horizon, as shown in Fig. \ref{fig:obs_horizon_ablation}. However, vision-based Diffusion Policy, in particular the variant with CNN backbone, see performance decrease with increasing observation horizon. In practice, we found an observation horizon of 2 is good for most of the tasks for both state and image observations.

\subsection{Performance Improvement Calculation}
For each task $i$ (column) reported in Tab. \ref{tab:table_low_dim}, Tab. \ref{tab:table_image} and Tab. \ref{tab:multi_stage} (mh results ignored), we find the maximum performance for baseline methods $max\_baseline_i$ and the maximum performance for Diffusion Policy variant (CNN vs Transformer) $max\_ours_i$. For each task, the performance improvement is calculated as $improvement_i = \frac{max\_ours_i-max\_baseline_i}{max\_baseline_i}$ (positive for all tasks). Finally, the average improvement is calculated as $avg\_improvement=\frac{1}{N}\sum^i_N improvement_i=0.46858 \approx 46.9\%$.

\section{Realworld Task Details}

\subsection{Push-T}
\subsubsection{Demonstrations}
136 demonstrations are collected and used for training. The initial condition is varied by randomly pushing or tossing the T block onto the table. Prior to this data collection session, the operator has performed this task for many hours and should be considered proficient at this task.

\subsubsection{Evaluation}
We used a fixed training time of 12 hours for each method, and selected the last checkpoint for each, with the exception of IBC, where the checkpoint with minimum training set action prediction MSE error due to IBC's training stability issue. The difficulty of training and checkpoint selection for IBC is demonstrated in main text Fig. 7.
Each method is evaluated for 20 episodes, all starting from the same set of initial conditions. To ensure the consistency of initial conditions, we carefully adjusted the pose of the T block and the robot according to overlayed images from the top-down camera.
Each evaluation episode is terminated by either keeping the end-effector within the end-zone for more than 0.5 second, or by reaching the 60 sec time limit.
The IoU metric is directly computed in the top-down camera pixel space.

\subsection{Sauce Pouring and Spreading}
\subsubsection{Demonstrations} 
50 demonstrations are collected, and 90\% are used for training for each task. 
For pouring, initial locations of the pizza dough and sauce bowl are varied. After each demonstration, sauce is poured back into the bowl, and the dough is wiped clean.
For spreading, location of the pizza dough as well as the poured sauce shape are varied. For resetting, we manually gather sauce towards the center of the dough, and wipe the remaining dough clean. 
The rotational components for tele-op commands are discarded during  spreading and sauce transferring to avoid accidentally scooping or spilling sauce.

\subsubsection{Evaluation}
Both Diffusion Policy and LSTM-GMM are trained for 1000 epochs. The last checkpoint is used for evaluation.

Each method is evaluated from the same set of random initial conditions, where positions of the pizza dough and sauce bowl are varied. We use a similar protocol as in \textbf{Push-T} to set up initial conditions. We do not try to match initial shape of poured sauce for spreading. Instead, we make sure the amount of sauce is fixed during all experiments. 

The evaluation episodes are terminated by moving the spoon upward (away form the dough) for 0.5 seconds, or when the operator deems the policy's behavior is unsafe.

The coverage metric is computed by first projecting the RGB image from both the left and right cameras onto the table space through homography, then computing the coverage in each projected image. The maximum coverage between the left and right cameras is reported.

\section{Realworld Setup Details}

\subsubsection{UR5 robot station}
\label{sec:ur5_setup}
Experiments for the \textbf{Push-T} task are performed on the UR5 robot station. 

The UR5 robot accepts end-effector space positional command at 125Hz, which is linearly interpolated from the 10Hz command from either human demonstration or the policy. The interpolation controller limits the end-effector velocity to be below 0.43 m/s and its position to be within the region 1cm above the table for safety reason. Position-controlled policies directly predicts the desired end-effector pose, while velocity-controlled policies predicts the difference the current positional setpoint and the previous setpoint. 

The UR5 robot station has 5 realsense D415 depth camera recording 720p RGB videos at 30fps. Only 2 of the cameras are used for policy observation, which are down-sampled to 320x240 at 10fps.

During demonstration, the operator teleoperates the robot via a 3dconnexion SpaceMouse at 10Hz.

\subsection{Franka Robot Station}
\label{sec:franka_setup}
Experiments for \textbf{Sauce Pouring and Spreading, Bimanual Egg Beater, Bimanual Mat Unrolling, and
Bimanual Shirt Folding} tasks are performed on the Franka robot station. 

For the non-haptic control, a custom mid-level controller is implemented to generate desired joint positions from desired end effector poses from the learned policies. At each time step, we solve a differential kinematics problem (formulated as a Quadratic Program) to compute the desired joint velocity to track the desired end effector velocity. The resulting joint velocity is Euler integrated into joint position, which is tracked by a joint-level controller on the robot. This formulation allows us to impose constraints such as collision avoidance for the two arms and the table, safety region for end effector and joint limits. It also enables regulating redundant DoF in the null space of the end effector commands. This mid-level controller is particularly valuable for safeguarding the learned policy during hardware deployment. 
% On the other hand, since the optimization is greedy per time step, naively commanding large end-effector displacements can easily drive the QP into a corner on the constraint manifold.

For haptic teleoperation control, another custom mid-level controller is implemented, but formulated as a pure torque-controller. The controller is formulated using Operational Space Control \citet{khatib1987osc} as a Quadratic Program operating at 200 Hz, where position, velocity, and torque limits are added as constraints, and the primary spatial objective and secondary null-space posture objectives are posed as costs.
This, coupled with a good model of the Franka Panda arm, including reflected rotor inertias, allows us to perform
good tracking with pure spatial feedback, and even better tracking with feedforward spatial acceleration.
Collision avoidance has not yet been enabled for this control mode.

Note that for inference, we use the non-haptic control. Future work intends to simplify
this control strategy and only use a single controller for our given objectives.

The operator uses a SpaceMouse or VR controller input device(s) to control the robot's end effector(s), and the grippers are controlled by a trigger button on the respective device. Tele-op and learned policies run at 10Hz, and the mid-level controller runs around 1kHz. Desired end effector pose commands are interpolated by the mid-level controller. This station has 2 realsense D415 RGBD camera streaming VGA RGB images at 30fps, which are downsampled to 320x240 at 10fps as input to the learned policies.

\subsection{Initial and Final States of Bimanual Tasks} 
\label{sec:bimanual_ini_fial}
The following figures show the initial and final state of four bimanual tasks. Green and red boxes indicate successful and failed rollouts respectively. Since mat and shirt are very flat objects, we used a homographic projection to better visualize the initial and final states.

\begin{figure}
    \centering
    \includegraphics[width=\linewidth]{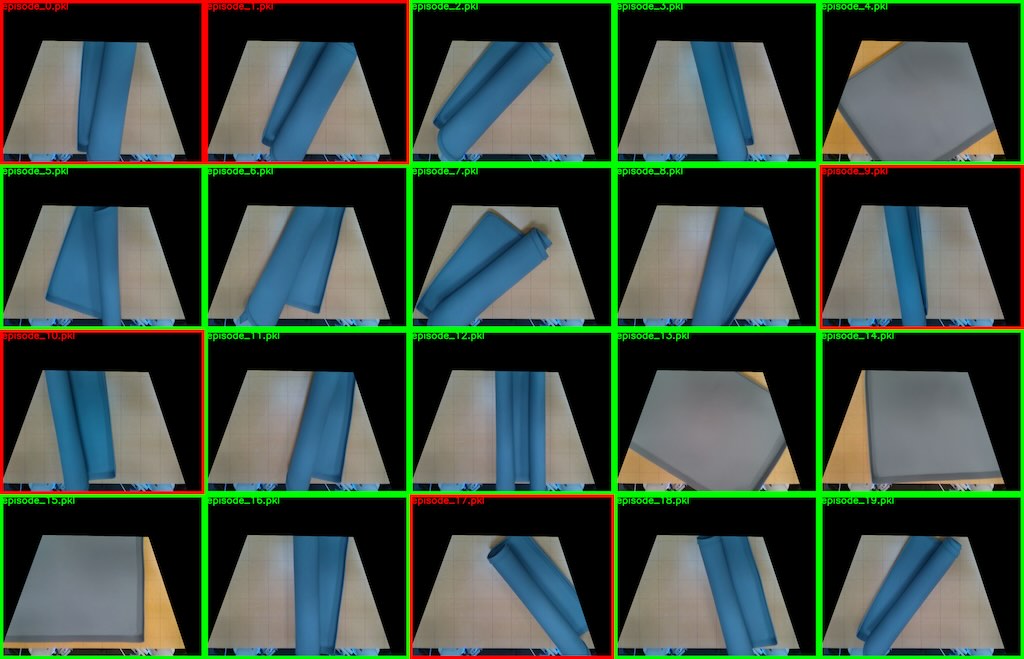}
    \caption{Initial states for Mat Unrolling}
    \label{fig:unroll_mat_ini}
% \end{figure}
% \begin{figure}
    \centering
    \includegraphics[width=\linewidth]{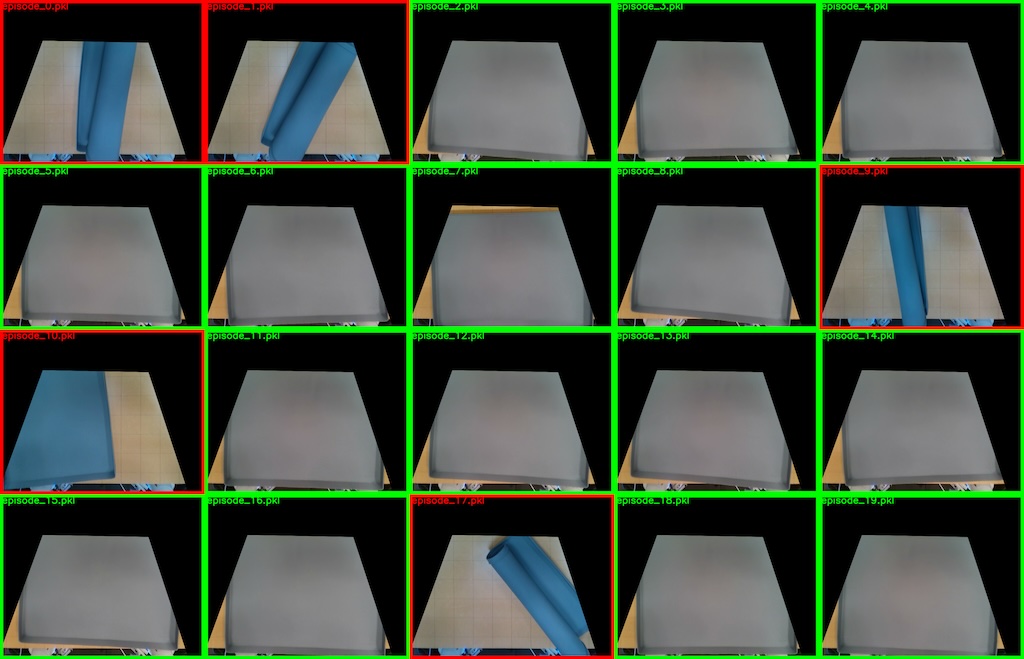}
    \caption{Final states for Mat Unrolling}
    \label{fig:unroll_mat_last}
\end{figure}

\begin{figure}
    \centering
    \includegraphics[width=\linewidth]{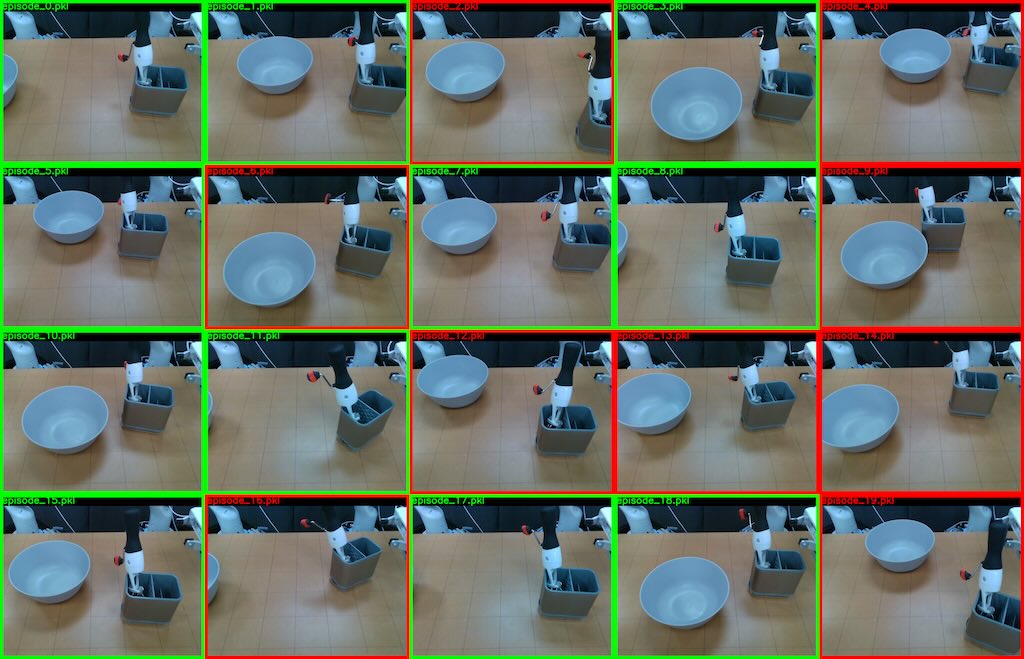}
    \caption{Initial states for Egg Beater}
    \label{fig:egg_beater_ini}
% \end{figure}
% \begin{figure}
    \centering
    \includegraphics[width=\linewidth]{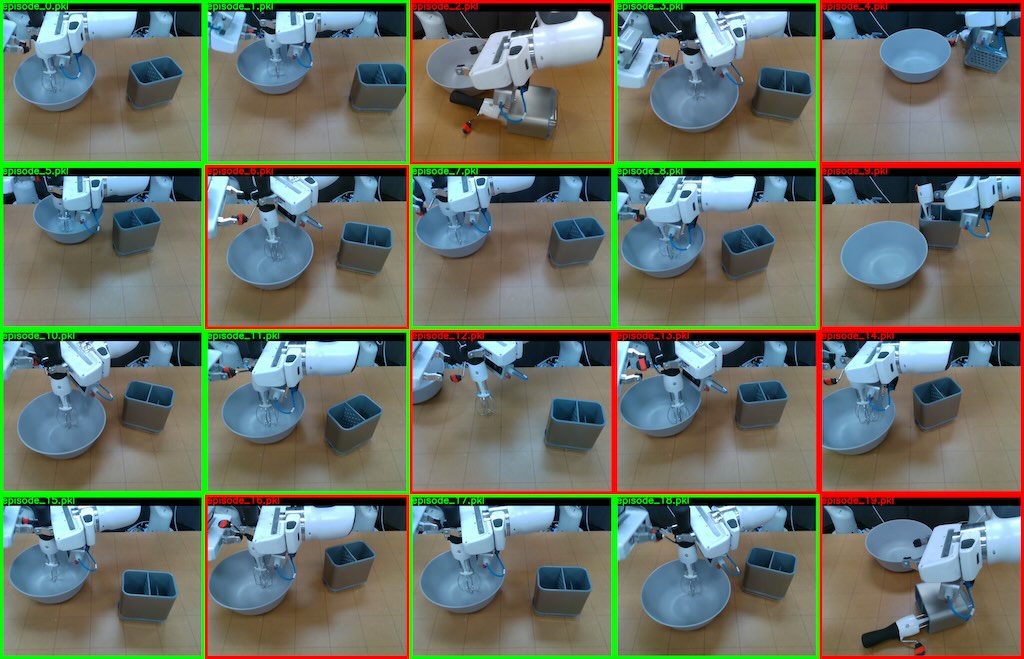}
    \caption{Final states for Egg Beater}
    \label{fig:egg_beater_last}
\end{figure}

\begin{figure}
    \centering
    \includegraphics[width=\linewidth]{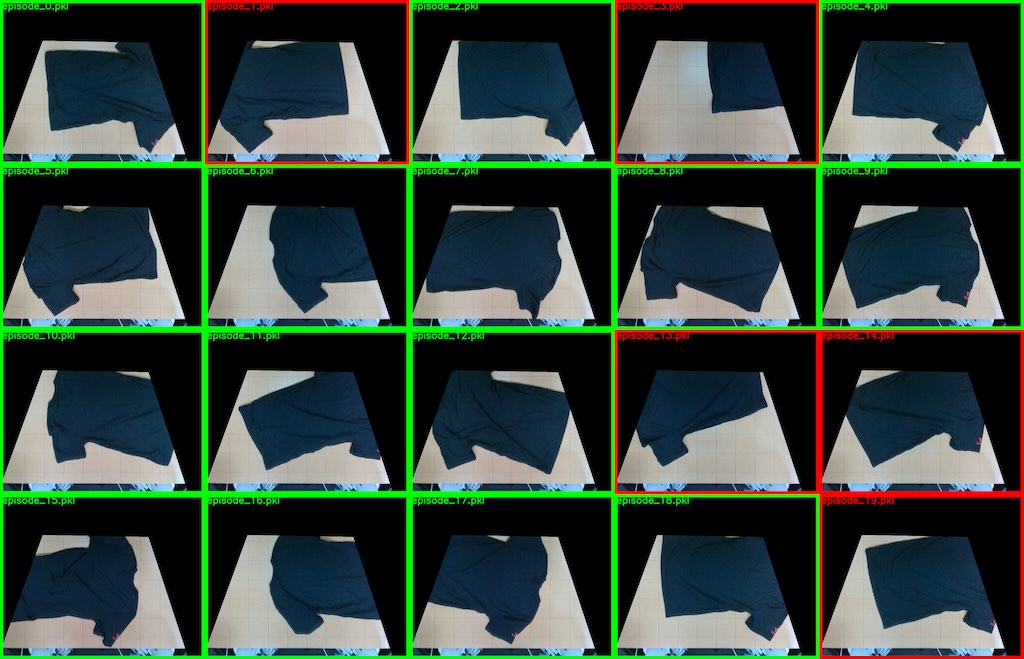}
    \caption{Initial states for Shirt Folding}
    \label{fig:fold_shirt_ini}
% \end{figure}
% \begin{figure}
    \centering
    \includegraphics[width=\linewidth]{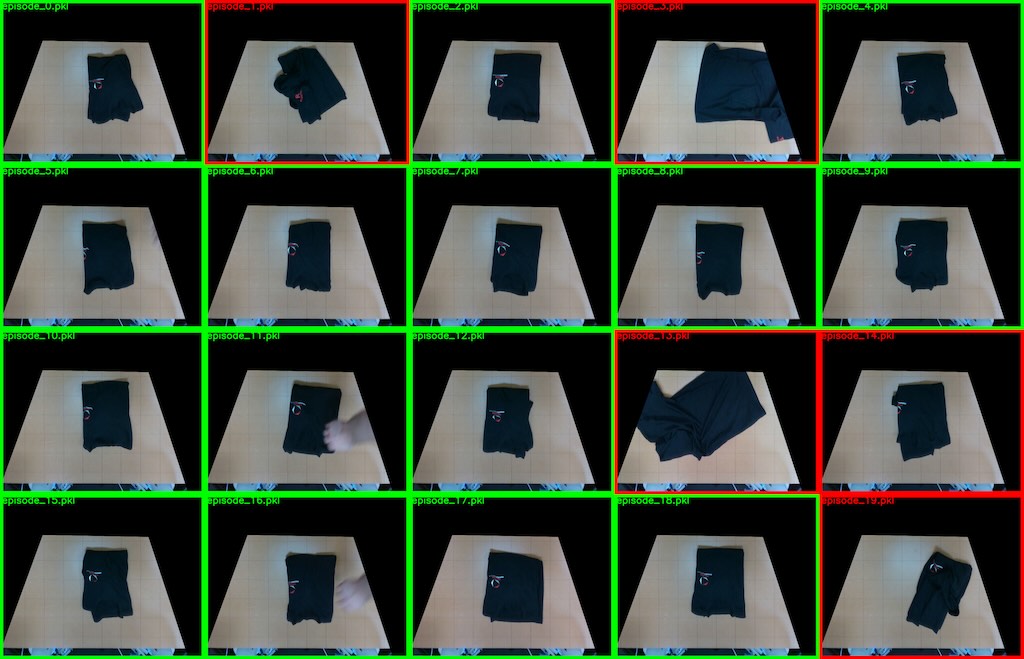}
    \caption{Final states for Shirt Folding}
    \label{fig:fold_shirt_last}
\end{figure}

\end{document}